\documentclass{article}
\usepackage{arxiv,times}
\usepackage{amsmath,amsfonts,bm}

\def\eqref#1{equation~\ref{#1}}
\def\1{\bm{1}}

\DeclareMathAlphabet{\mathsfit}{\encodingdefault}{\sfdefault}{m}{sl}
\SetMathAlphabet{\mathsfit}{bold}{\encodingdefault}{\sfdefault}{bx}{n}

\usepackage{hyperref}
\usepackage{url}
\usepackage{subcaption} 
\usepackage{wrapfig}
\usepackage[section]{placeins}
\usepackage{tabularx,ragged2e}

\usepackage{array,colortbl,xcolor,makecell,graphicx,booktabs}
\usepackage{xspace}
\usepackage{amsthm}
\usepackage[most]{tcolorbox}
\usepackage{enumitem}
\usepackage{tikz}
\usepackage{fontawesome5,twemojis}
\usepackage{microtype}
\usepackage{dsfont}

\newcolumntype{Y}{>{\RaggedRight\arraybackslash}X}

\definecolor{secgray}{RGB}{246,246,246}
\definecolor{routebg}{RGB}{244,248,255}
\definecolor{surveybg}{RGB}{255,249,238}
\definecolor{statbg}{RGB}{231,243,255}
\definecolor{dynbg}{RGB}{255,239,231}

\definecolor{corrbg}{RGB}{231,243,255}
\definecolor{errbg}{RGB}{238,255,240}
\definecolor{stabg}{RGB}{255,245,230}
\definecolor{updbg}{RGB}{244,236,255}
\definecolor{uncbg}{RGB}{255,235,238}

\theoremstyle{definition}

\tcolorboxenvironment{definition}{
  enhanced,
  breakable,
  colback=white,
  colframe=black!60,
  boxrule=0.6pt,
  arc=1.2pt,
  left=8pt,right=8pt,top=6pt,bottom=6pt,
  before skip=10pt plus 2pt minus 2pt,
  after skip=10pt plus 2pt minus 2pt,
}

\setlist[description]{leftmargin=1.8em,labelsep=0.6em,font=\normalfont\bfseries}

\setlist[description]{leftmargin=1.8em,labelsep=0.6em,font=\normalfont\bfseries}

\newlength{\rescolw}
\newlength{\rescolwexp}
\newcolumntype{C}[1]{>{\centering\arraybackslash}m{#1}}
\newcommand{\roth}[1]{\makebox[\rescolw][c]{\rotatebox[origin=c]{25}{\footnotesize\strut #1}}}

\definecolor{keyblue}{HTML}{5A85E1}
\definecolor{keybg}{HTML}{F1F7FF}

\newtcolorbox{keytakeaways}[1]{
  enhanced,
  breakable,
  colback=keybg,
  colframe=keyblue,
  boxrule=0pt,
  arc=8pt,
  left=14pt, right=14pt, top=11pt, bottom=8pt,
  borderline west={4pt}{0pt}{keyblue},
  before skip=10pt, after skip=10pt,
  overlay={
    \node[
      fill=keyblue,
      text=white,
      font=\bfseries,
      rounded corners=5pt,
      inner xsep=10pt,
      inner ysep=4pt,
      anchor=west
    ] at ([xshift=10pt,yshift=-1pt]frame.north west)
    {\faLightbulb\hspace{6pt}Key Findings: #1};
  },
}

\definecolor{defblue}{HTML}{7883B9}
\definecolor{defbg}{HTML}{F7FBFF}

\newtcolorbox{definitionbox}[1]{%
  enhanced,
  breakable,
  colback=defbg,
  colframe=defblue,
  boxrule=0pt,
  arc=8pt,
  left=14pt, right=14pt, top=11pt, bottom=8pt,
  borderline west={4pt}{0pt}{defblue},
  before skip=10pt, after skip=10pt,
  overlay={
    \node[
      fill=defblue,
      text=white,
      font=\bfseries,
      rounded corners=5pt,
      inner xsep=10pt,
      inner ysep=4pt,
      anchor=west
    ] at ([xshift=10pt,yshift=-1pt]frame.north west)
    {\faBook\hspace{6pt}Definition: #1};
  },
}

\setlist[itemize]{leftmargin=1.4em, itemsep=6pt, topsep=4pt}

\title{\large\raisebox{-0.2\height}{\includegraphics[height=3.5em]{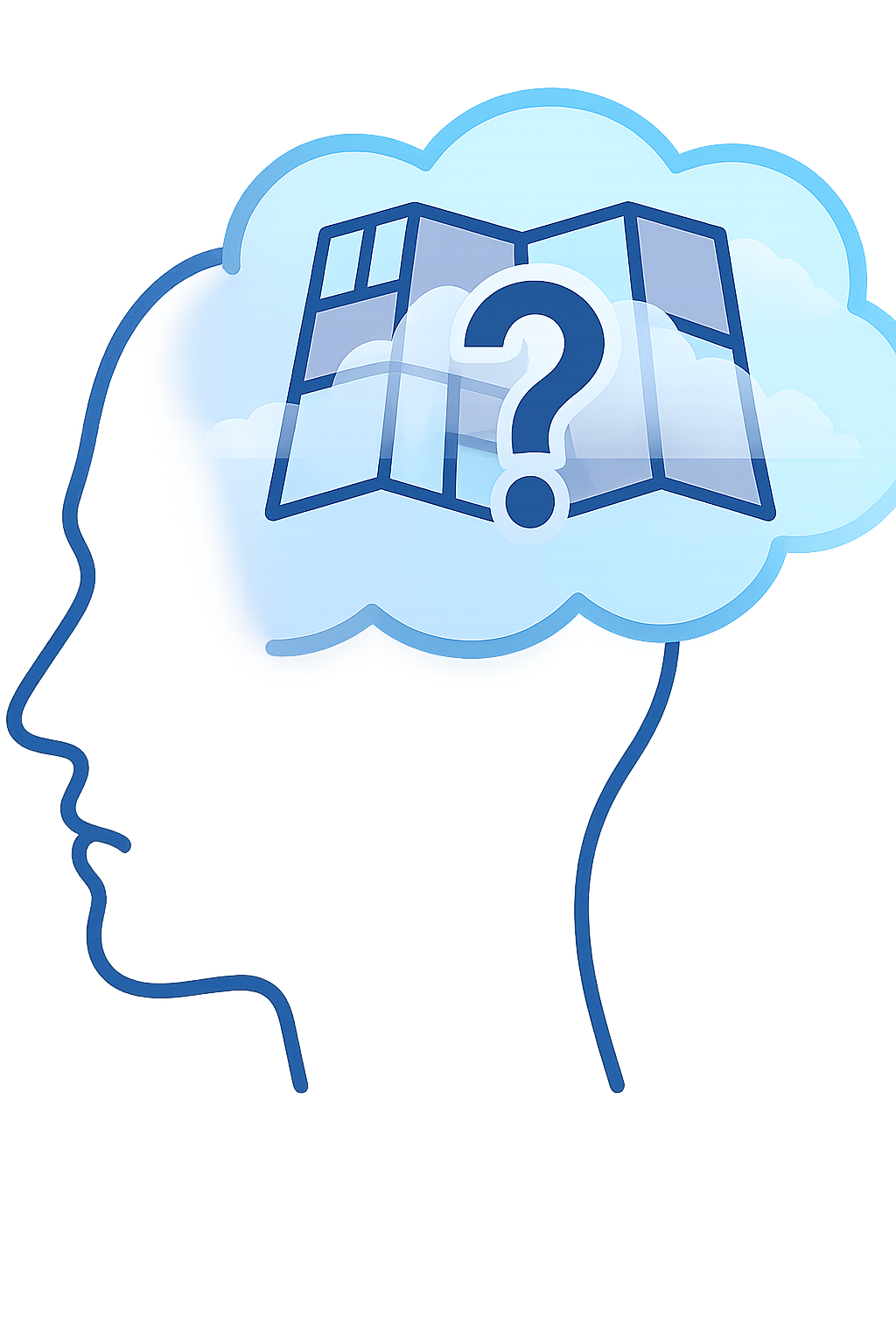}}\!
\fontsize{15}{18}\selectfont
Theory of Space: Can Foundation Models Construct Spatial Beliefs through Active Exploration?
}

\author{
\vspace{3em}
\begin{tabular}{c}
Pingyue Zhang$^{1,*,\dagger}$, Zihan Huang$^{*}$, Yue Wang$^{4,*}$, Jieyu Zhang$^{3,*}$, \\
Letian Xue$^{1}$, Zihan Wang$^{1}$, Qineng Wang$^{1}$, Keshigeyan Chandrasegaran$^{2}$, \\
Ruohan Zhang$^{2}$, Yejin Choi$^{2}$, Ranjay Krishna$^{3}$, Jiajun Wu$^{2}$, Li Fei-Fei$^{2}$, Manling Li$^{1,\dagger}$ \\
\\
\normalfont\normalsize
$^{1}$Northwestern University \quad
$^{2}$Stanford University  \quad
$^{3}$University of Washington \quad
$^{4}$Cornell University \\
\normalfont\normalsize
\texttt{pingyuezhang@u.northwestern.edu, manling.li@northwestern.edu} \\
\\
\normalfont\normalsize
{\small $^{*}$Equal contribution \quad $^{\dagger}$Corresponding author}
\end{tabular}
\vspace{-30pt}
}

\DeclareRobustCommand{\tos}{\textsc{Theory of Space}\xspace}

\iclrfinalcopy
\begin{document}

\maketitle

\begin{abstract}

Spatial embodied intelligence often operates under partial observability, where agents must act to acquire missing information rather than passively consume complete observations.
In such settings, progress depends on actively selecting informative actions that reduce uncertainty and support the construction of spatial understanding.
While multimodal foundation models have shown strong performance on passive multimodal perception and reasoning tasks, their ability to support active, self-directed exploration under partial observability has not been systematically studied.
In particular, it remains unclear whether and how these models can decide what to observe next in order to build and maintain a coherent spatial belief over time.
We therefore propose \tos, defined as an agent’s ability to actively acquire information through self-directed, active exploration and to construct, revise, and exploit a spatial belief from sequential, partial observations.
We implement \tos using a benchmark with textual and visual environments. Rather than solving specific tasks, the goal is curiosity-driven exploration to build a complete, accurate spatial belief. 
A core innovation is spatial belief probing: we prompt it to reveal its internal spatial belief as a cognitive map at each step, letting us measure the quality of its underlying spatial model.
Our evaluation of state-of-the-art models on a suite of downstream tasks reveals critical bottlenecks: (1) \textbf{The Active-Passive Gap}: Performance degrades when agents must autonomously gather information (e.g., \textsc{GPT-5.2}: $0.57{\to}0.46$); (2) \textbf{Inefficiency}: Models explore in an unsystematic way and with high redundancy, failing to match the efficiency of program-based proxies while producing no better results.
Through belief probing, we diagnose that perception acts as an initial bottleneck, yet global beliefs suffer further from \textbf{instability} that causes spatial knowledge to degrade over time. Finally, using a false belief paradigm to test belief revision, we uncover \textbf{Belief Inertia} where agents fail to overwrite obsolete priors. This issue exists in text agents but is notably severe in vision-based models. 

\end{abstract}

\vspace{-2em}
\noindent
\setlength{\tabcolsep}{0pt}
\renewcommand{\arraystretch}{1.15}

\begin{center}
\begin{minipage}{0.95\linewidth}
\raggedright
\begin{tabularx}{\linewidth}{@{}l@{\hspace{0.8em}}l@{\hspace{1.2em}}>{\raggedright\arraybackslash}X@{}}

\faGlobe\! &
\textbf{Website} &
\href{https://theory-of-space.github.io/}{\footnotesize\texttt{https://theory-of-space.github.io/}} \\

\faGithub\! &
\textbf{Code} &
\href{https://github.com/mll-lab-nu/Theory-of-Space}{\footnotesize\texttt{https://github.com/mll-lab-nu/Theory-of-Space}} \\

\faDatabase\! &
\textbf{Data} &
\href{https://huggingface.co/datasets/MLL-Lab/tos-data}{\footnotesize\texttt{https://huggingface.co/datasets/MLL-Lab/tos-data}} \\

\end{tabularx}
\end{minipage}
\end{center}

\vspace{-0.8em}

\section{Introduction}

\begin{figure}[!htbp]
  \centering
  \includegraphics[width=0.95\linewidth]{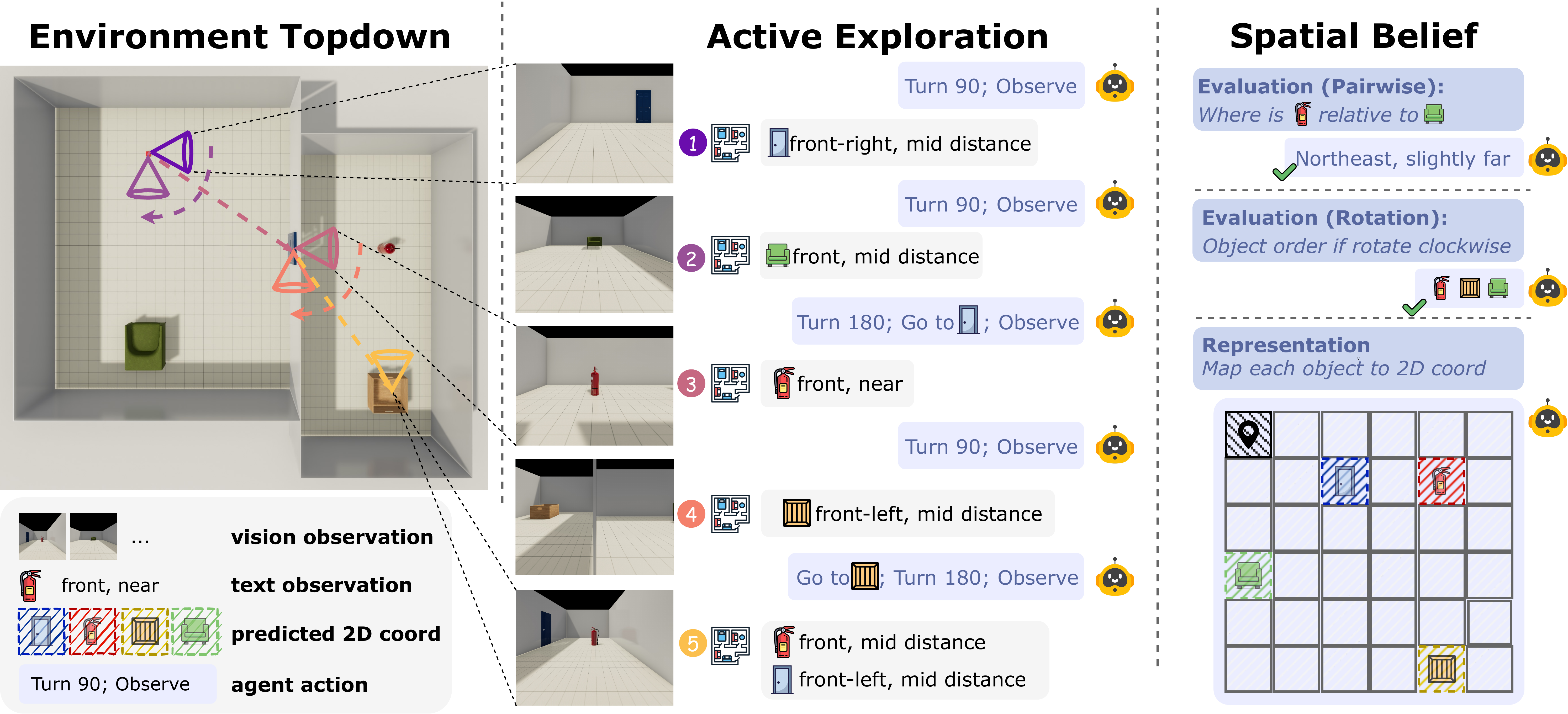}
  \caption{\textbf{\tos: active exploration, probed belief, and evaluation.} 
Left: a top-down view of agent trajectory under partial observability in multiple-room scenes. 
Middle: the agent’s action loop of moving, rotating, and observing in text- or vision-based environments, receiving egocentric observations and updating an internal belief. 
Right: evaluation through exploitation of the belief in spatial tasks and direct probing via probed cognitive maps.}
  \vspace{-10pt}
  \label{fig:tos_final}
\end{figure}

Spatial embodied intelligence relies on active exploration. Unlike disembodied systems that passively process fixed observations, an embodied agent could take actions to alter its position in the environment as \emph{exploration}, selectively acquiring observations needed to construct spatial knowledge for various spatial tasks. 
Cognitive science shows that such active exploration leads to substantially better spatial understanding than passively receiving the same information, even when observations are identical~\citep{held1963movement,chrastil2012active, chrastil2013active}.
But exploration isn’t simply about collecting more observations. It is about efficiency, acting under uncertainty to target what is unknown or ambiguous in the agent’s spatial belief and maximize information gain.

We propose \tos as a framework that explicitly treats exploration as a first-class decision-making problem, decoupled from any single downstream task, focusing on opening the box of the agent’s internal spatial belief. 
Just as Theory of Mind (ToM) measures how agents model the hidden mental states of others, \tos assesses an agent's ability to model the hidden physical structure of the world. 
We define \tos as an embodied agent’s ability to actively \textbf{construct}, \textbf{revise} in a dynamic environment, and \textbf{exploit} an internal \emph{spatial belief} formed through active exploration.
Beyond end-task evaluation, \tos directly probes what the agent knows, what remains uncertain, and how effectively its actions reduce those uncertainties, measured by the number of exploration steps and the uncertainty resolved per action. Figure~\ref{fig:tos_final} provides an overview of \tos's active exploration, belief probing, and end-task evaluation.

We apply \tos to evaluate multimodal language models, which are promising candidates for embodied agents. By integrating vision and language, they support unified perception, reasoning, and action over time, yet existing foundation-model benchmarks offer little insight into these capabilities.
Most current benchmarks fall into two categories: \textit{passive} \citep{weston2015towards, shi2022stepgame, yang2025mmsi, gholami2025spatial, yang2025thinking}, where the agent is only asked to reason over given observations, and \textit{task-driven} \citep{gordon2018iqa, shridhar2020alfred, li2025embodiedagentinterfacebenchmarking, yang2025embodiedbenchcomprehensivebenchmarkingmultimodal}, where the agent must achieve a specific goal (e.g., ``\textit{find the red chair}'').

\begin{wrapfigure}{r}{0.56\textwidth}
  \centering
  \vspace{-0.5\baselineskip}
  \includegraphics[width=0.52\textwidth]{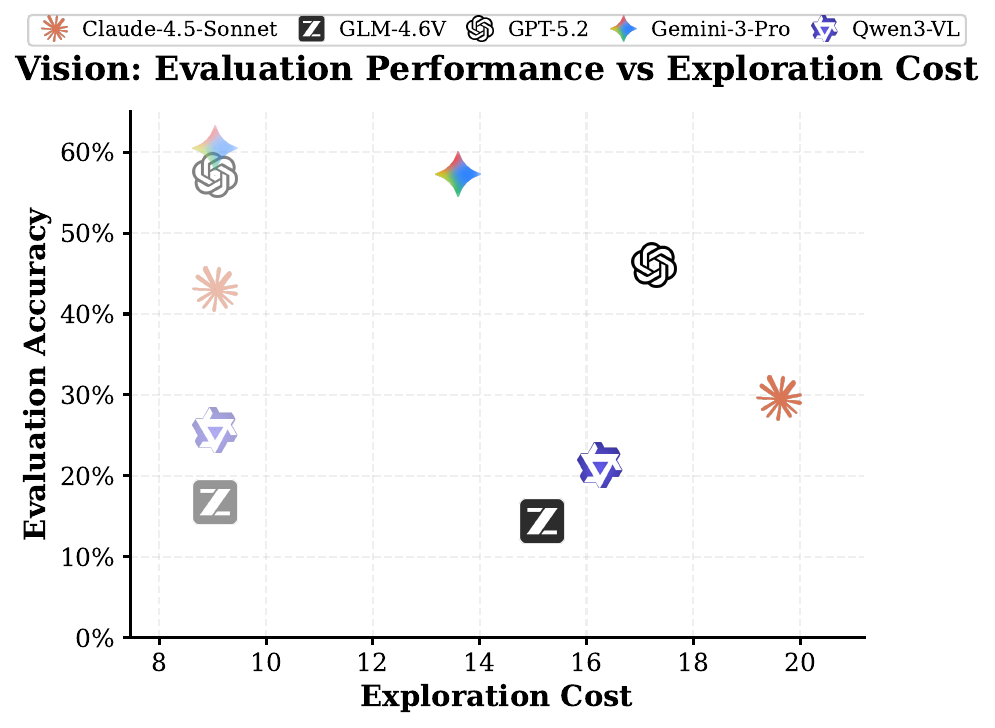}
  \caption{\textbf{Evaluation accuracy vs. exploration cost for active exploration in vision-world.} Faded icons mark the passive setting, where the agent gets a pre-generated exploration history and only reasons.}
  \label{fig:exp_vs_eval}
  \vspace{-0.5\baselineskip}
\end{wrapfigure}

In this work, we propose to systematically evaluate the active process of spatial belief construction.  
Unlike passive benchmarks, our \tos benchmark requires agents to actively explore via \textit{moving}, \textit{rotating}, and \textit{observing} to build coherent global beliefs. We implement a scalable environment using ThreeDWorld~\citep{gan2021threedworldplatforminteractivemultimodal} and Objaverse~\citep{deitke2022objaverseuniverseannotated3d} that provides \textit{Text-based} and \textit{Vision-based} worlds to localize perception versus reasoning failures.
After active exploration, we evaluate the process along two axes: (i) \textbf{belief exploitation} via spatial downstream tasks that probe \textit{route}-level and \textit{survey}-level knowledge~\citep{siegel1975development,montello1998framework}; and (ii) \textbf{exploration efficiency} via the number of exploration steps and the accumulated information gain curve over steps, capturing how quickly an agent reduces uncertainty rather than merely increasing coverage. 
Finally, we design scripted proxy agents that execute strong reference trajectories to disentangle exploration from reasoning.
Our evaluation of state-of-the-art foundation models reveals both promising capability in the pure text world and striking limitations in the vision world under \tos. \textbf{Active exploration remains a primary bottleneck.} Models perform reasonablely well in passive setting, but degrade when they must actively gather information (e.g., \textsc{GPT-5.2}: $57.1 \rightarrow 46.0$; \textsc{GEMINI-3 PRO}: $60.5 \rightarrow 57.3$; Figure.~\ref{fig:exp_vs_eval}).
We also find a major efficiency gap: rule-based proxy agents reach target coverage in $\sim 9$ steps, whereas foundation models explore redundantly, requiring $\geq 14$ steps without improving belief accuracy. Thus, even when models can reason about spatial tasks (as reflected in passive performance), they fail to autonomously structure the information-gathering needed to solve them.

Beyond downstream task scores, a core contribution of \tos is \textbf{explicit cognitive-map probing}, which provides a direct window into the agent’s latent spatial belief as it is constructed and revised. Rather than treating the agent as a black box whose internal state is only inferred from final answers, we prompt the model to expose its evolving cognitive map during exploration, enabling measurement of both belief accuracy and belief uncertainty at each step.
This probing-based assessment uniquely supports fine-grained diagnosis of \emph{how} models represent space: it reveals that while perception acts as an initial bottleneck, global beliefs also suffer severely from instability, causing knowledge to degrade over time.
This allows us to track belief evolution over time, attribute failures to specific representational breakdowns, and evaluate whether an agent truly “knows what is uncertain” rather than merely producing plausible outputs.

Finally, to evaluate the mechanics of dynamic spatial updating, we introduce a \emph{False Belief} paradigm. By altering the environment (relocating or reorienting objects) after the agent's initial exploration, we uncover a phenomenon we term \textbf{spatial belief inertia}: agents (particularly in vision-based settings) struggle to overwrite obsolete spatial priors with new sensory evidence. Despite directly observing the new configuration, models persist in their initial, now incorrect coordinates. This reveals a critical failure in spatial memory revision, where foundational models lack the plasticity to revise their internal cognitive maps in response to physical changes.

An important direction for future work is to extend \tos beyond single-agent settings to multi-agent exploration, where additional challenges arise around coordination and aligning (or sharing) spatial beliefs across agents.

\section{Theory of Space}
To build agents with spatial intelligence, we argue for evaluating not merely passive reasoning, but the \textbf{active, self-directed construction} of spatial belief from partial observations. We introduce \tos, a conceptual counterpart to \textit{Theory of Mind} (ToM).  While ToM models hidden mental states of others, \tos models \textbf{uncertain, currently unobserved} structure of space. 

\begin{definitionbox}{\tos}
Ability to \textbf{construct}, \textbf{revise}, and \textbf{exploit} an internal spatial belief.
\end{definitionbox}
Here, an \textbf{internal spatial belief} is a mental model~\citep{taylor1992spatial} of spatial layout and relations maintained in working memory and updated from partial observations.
% \tos is defined as an agent’s ability to perform three core functions with an internal spatial belief: (1) \textbf{construct} a globally consistent internal spatial belief by actively seeking out and integrating partial observations; (2) \textbf{revise} dynamically this internal belief by using new information acquired through further exploration to resolve conflicts with prior beliefs; (3) \textbf{exploit} the current belief to support spatial tasks and guide future exploration to reduce uncertainty. 
We formalize \tos within a partially observable framework over a spatial structure $S \in \mathcal{S}$. The agent interacts with $S$ to generate a history $h_t = (o_{0:t}, a_{0:t})$, where $o$ and $a$ denote observations and actions. We define \tos as the capacity to manipulate a probabilistic belief $B_t$ through three core operations:

\begin{enumerate}
    \item \textbf{Construct:} \emph{To form a globally consistent internal spatial belief by actively seeking out and integrating partial observations.} Formally, the agent integrates $h_t$ to approximate the true posterior, denoted as $B_t(S) \approx P(S \mid h_t)$.
    \item \textbf{Revise:} \emph{To dynamically update the internal belief by using new information acquired through further exploration to resolve conflicts with prior beliefs.} Upon an environmental shift $S \to S'$, the agent utilizes exploratory actions $\Delta h$ to minimize the divergence from the new ground truth, i.e., $B_{t+\Delta t} \to P(S' \mid h_{t+\Delta t})$.
    \item \textbf{Exploit:} \emph{To utilize the current belief to support spatial tasks.} The agent utilizes a policy $\pi$ conditioned on the belief, $\pi(a_t \mid B_t)$, to perform a downstream task $\mathcal{T}$. In a benchmark context, we measure the \textit{value of belief} by the performance metric $\mathcal{J}$ achieved by this policy: $\mathcal{J}(\pi(\cdot|B_t), \mathcal{T})$.
\end{enumerate}

\subsection{A Paradigm for assessing \tos of Large Foundation Models}
\label{sec:tos4llm}
We propose a new paradigm for Assessing \tos of large foundation models, which consists of three essential components below. 

\label{sec:spatial-tasks}

\textbf{Task-Agonistic Active Exploration to Move From Passive Viewer to Active Explorer.} 
Evaluating \tos requires a shift from downstream tasks to exploration, i.e., how an agent explores and decides ``what to see next''. 
In detail, we place the agent in a partially observable environment and explicitly challenge the LLM/VLM agent to actively select actions for itself, including \textit{moving}, \textit{rotating}, \textit{observing}, and \textit{terminating}. The primary goal is not to complete a downstream task or follow pre-collected trajectories, but to build a general-purpose internal model from its own self-directed exploration with minimal cost. 
This process encompasses both initial \textbf{Belief Construction} and dynamic \textbf{Belief Revision}. Inspired by the \textit{false belief} paradigm in Theory of Mind~\citep{wimmer1983belief} and \textit{spatial belief revision}~\citep{knauff2013spatial}, we evaluate whether an agent can detect dynamic environmental changes and correctly revise its internal belief during exploration. This demonstrates the ability to customize beliefs given evolving observations. 
Consequently, the model must identify what remains uncertain and actively terminate exploration only upon acquiring sufficient evidence to form an accurate and responsive internal map.

\textbf{Belief Exploitation Assessment.}
To translate \tos into concrete evaluation tasks, we draw insights from the development of spatial representations ~\citep{siegel1975development,montello1998framework} 
and define two tasks to measure an agent's ability to exploit its internal belief for goal-directed behavior:  (1) \textbf{Belief on Route} evaluates a path-based understanding of space organized around landmarks such as pairwise spatial relationships along a \textit{egocentric} navigation; (2) \textbf{Belief on Survey} assesses  a map-like ``bird's-eye view'' that represents space allocentrically, allowing for the inference of global relationships.

\textbf{Explicit Probing of the Internal Spatial Belief. } 
Behavioral success such as whether the agent finds the chair cannot directly reveal the quality of agent’s internal model. We require the agent to explicitly represent its spatial belief by \textit{probing its cognitive map} at any point of exploration. 
Cognitive maps are structured allocentric representations of space, which is well-established in neuroscience~\citep{tolman1948cognitive,o1971hippocampus, hafting2005microstructure}. 
Thus, we use cognitive maps as the canonical representation of the hidden structure of space. 
In our implementation, we probe the agent's internal belief by requiring it to externalize a structured cognitive map. We evaluate the map’s \textbf{Correctness}, and we diagnose reasoning breakdowns with dynamic signals that capture how reliably observations are integrated, tracked over time, and kept coherent across local and global structure. 
Additionally, we explicitly test the agent's belief on uncertainty by identifying unobserved regions to measure its uncertainty modeling.
This shifts the evaluation from behavioral success to a direct assessment of representational competence, giving us a window into the agent's spatial belief development. 

\section{Benchmarking Theory of Space Ability for Foundation Models}

\label{sec:benchmark-pipeline}
Unlike task-driven benchmarks that only test task completion, we aim to answer ``\emph{can the agent form a global environmental belief through exploration?}''. We structure the benchmarking into two phases. In the \textbf{Exploration Phase I}, the agent interacts with the environment to construct spatial belief by selecting and executing actions in the action space in $\S$~\ref{sec:benchmark-env}, and gather a sequence of local observations to integrate them into a unified spatial belief. 
In the \textbf{Reasoning Phase II}, the agent is asked to conduct spatial tasks (detailed in $\S$~\ref{sec:spatial-tasks-detail}).

\subsection{Spatial Environment Construction}
\label{sec:benchmark-env}

To ensure controlled experimentation, we procedurally generate multi-room indoor layouts on an $N \times M$ grid. Each scene is populated with $n$ indoor objects, each assigned a 2D integer coordinate and a cardinal orientation from (\texttt{N}, \texttt{S}, \texttt{E}, \texttt{W}). 
The agent begins at a random position, is informed of the total number of rooms and the names of all objects in the scene, and then starts exploration. 
Following the Gym-style interface~\citep{brockman2016openai}, we define procedurally generated, highly scalable environments in which each random seed deterministically instantiates a distinct multi-room layout. 

\textbf{Action Space in the Environment.}~~ The agent's interaction with the world is designed to focus on high-level decision-making rather than low-level motor control: 
\texttt{Goto} to move directly to a currently visible object;  
\texttt{Rotate} to turn in place by $90^\circ, 180^\circ,$ or $270^\circ$;  
\texttt{Observe} to perceive visible objects in the $90^\circ$ field of view;
and 
\texttt{Query} to obtain a visible object’s absolute 2D coordinates.
We additionally assign costs of 1 to \texttt{Observe} and 2 to \texttt{Query}, encouraging \texttt{Query} to be used only when necessary to resolve ambiguity. However, across all models \texttt{Query} is invoked only rarely, so we restrict attention to \texttt{Observe} and measure exploration efficiency by step count instead of action cost.

\textbf{Observation Feedback from a Text-Vision Parallel Environment.}
We offer both text-based and vision-based environments, enabling diagnostic analysis of spatial reasoning. Each \texttt{Observe} action returns both textual and visual feedback from a $90^\circ$ field of view. The \textbf{Text World} provides symbolic observations with discrete bins for direction and distance (e.g., “chair is front-left and near”, detailed below), isolating pure spatial reasoning. The \textbf{Visual World} instead supplies ego-centric RGB images rendered in ThreeDWorld~\citep{gan2021threedworldplatforminteractivemultimodal} with Objaverse assets~\citep{deitke2022objaverseuniverseannotated3d}, requiring perception to recover spatial relations. 
To calibrate perception in the visual setting, we provide two reference images, indicating unit distance ($1$ grid unit) / angle (a $22.5^\circ$ angular cone), and showing all objects with their names and canonical ``front'' orientation, respectively. Details are shown in Appendix \P~\ref{sec:app:vision-world}

\textbf{Spatial Relation Representation.}
\label{subsec:spatial_relation} 
To ensure that agents perceive and communicate about space using a consistent language across tasks and modalities, we discretize spatial relationships for directions and distances. 
For \textbf{allocentric direction}, we discretize into eight $45^\circ$ bins aligned with the four cardinal and four intercardinal directions, denoted compactly as $\{\texttt{N}, \texttt{NE}, \texttt{E}, \texttt{SE}, \texttt{S}, \texttt{SW}, \texttt{W}, \texttt{NW}\}$. Each bin spans $45^\circ$ around its heading (e.g., $\texttt{N}=[-22.5^\circ,22.5^\circ)$). 
For \textbf{egocentric direction}, within a $90^\circ$ forward field of view (FOV), we use five labels: \texttt{front-left} $[-45^\circ,-22.5^\circ)$, \texttt{front-slight-left} $[-22.5^\circ,0)$, \texttt{front} $0^\circ$, \texttt{front-slight-right} $(0,22.5^\circ]$, and \texttt{front-right} $(22.5^\circ,45^\circ]$. 
For \textbf{distance}, measured in map units independent of direction, we define six bins: \textit{same} $=0$, \textit{near} $(0,2]$, \textit{mid} $(2,4]$, \textit{slightly far} $(4,8]$, \textit{far} $(8,16]$, and \textit{very far} $(16,32]$. 

\subsection{Downstream Spatial Tasks}
\label{sec:spatial-tasks-detail}
 
\begin{figure}[t]
  \centering
    \centering
    \includegraphics[width=0.95\linewidth]{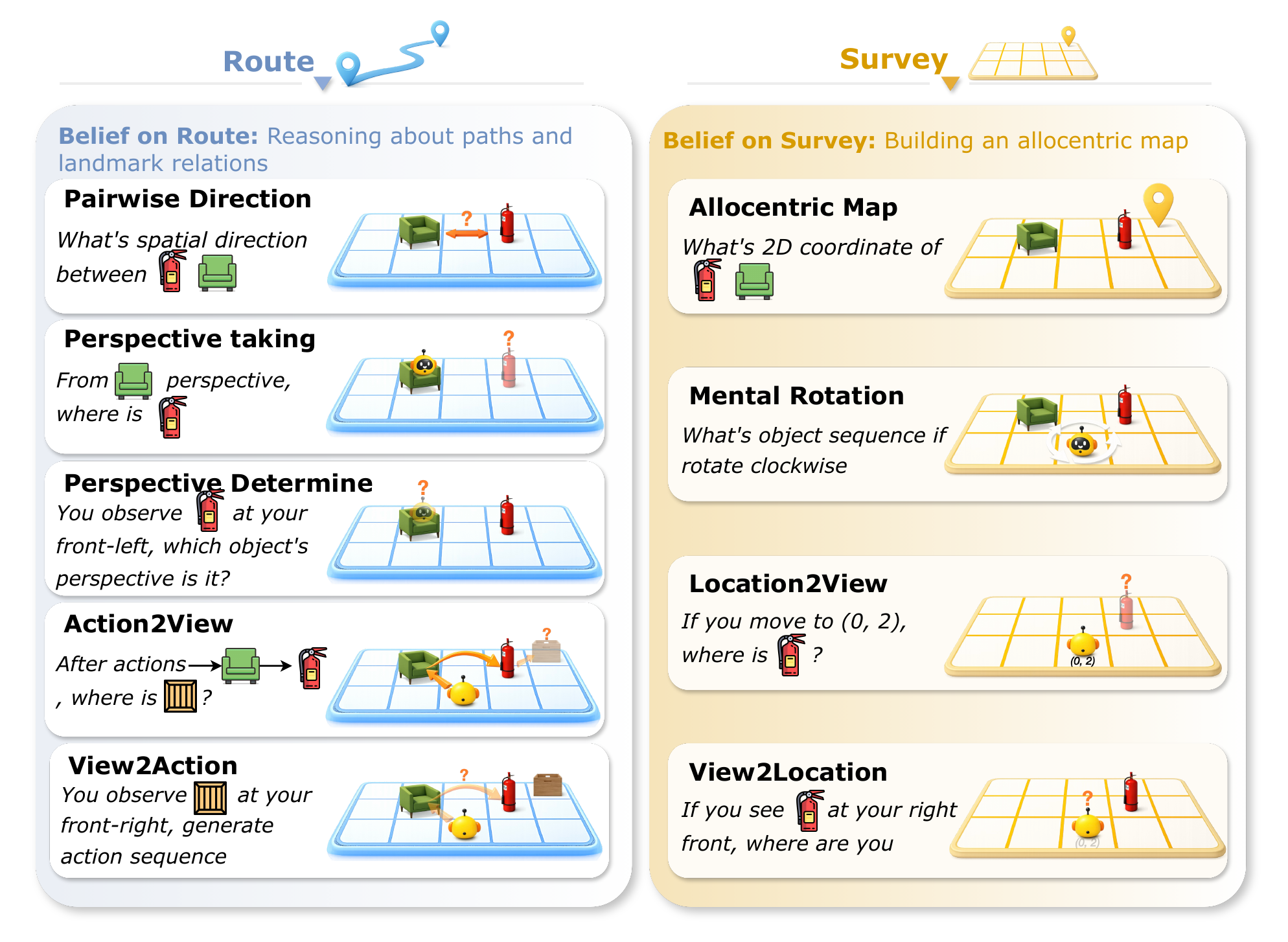}
    \caption{\textbf{\tos exploitation task suite:} it covers \textbf{route}-level egocentric reasoning and \textbf{survey}-level allocentric mapping. Route tasks evaluate path-based inference and egocentric observations. Survey tasks test global mapping, geometric transformation, and perspective conversion. Together they cover both local navigation reasoning and global spatial abstraction.}
    \label{fig:tos_eval}

\end{figure}

We use open-ended questions rather than multiple-choice questions to reduce the risk of knowledge leakage. Drawing on prior work~\citep{siegel1975development,montello1998framework}, we define tasks to evaluate an agent’s \textbf{Route} and \textbf{Survey} knowledge, shown in Table~\ref{tab:route-survey-comparison}.
Route belief captures how an agent encodes paths and spatial relations from an egocentric step-by-step perspective. Survey belief is a map-like, allocentric representation. 
An overview of the tasks is present in Figure~\ref{fig:tos_eval}.

\begin{table}[t]
\centering
\small
\setlength{\tabcolsep}{5pt}
\renewcommand{\arraystretch}{1.25}

\begin{tabularx}{\linewidth}{@{}>{\RaggedRight\arraybackslash}p{2.6cm}
@{\hspace{6pt}}Y
@{\hspace{10pt}}Y@{}}
\toprule
\textbf{Dynamic Group} & \textbf{Belief on Route} & \textbf{Belief on Survey} \\
\midrule

\textbf{Static} &
\textbf{Pairwise Relation (\textit{direction})}\par
report allocentric direction and distance from $A$ to $B$. &
\textbf{Allocentric Mapping (\textit{alloc.map})}\par
predict global coordinates (and headings) for all objects. \\
\midrule

\textbf{Forward Dynamics} &
\textbf{Perspective Taking (\textit{persp.take})}\par
output the observation from a specified object's perspective.\par\vspace{2pt}
\textbf{Action-to-View (\textit{act2view})}\par
given a sequence of \texttt{Goto}/\texttt{Rotate},
predict the final observation (one object in FOV with ego direction/distance bins). &
\textbf{Mental Rotation (\textit{ment.rot})}\par
predict the sequence of front-facing objects during a $360^\circ$ self-rotation.\par\vspace{2pt}
\textbf{Location2View (\textit{loc2view})}\par
given a global pose, predict the observation (one object in FOV with ego bins/distances). \\
\midrule

\textbf{Backward Dynamics} &
\textbf{Perspective Decision (\textit{perc.dec})}\par
infer which object's perspective the agent is currently adopting.\par\vspace{2pt}
\textbf{View-to-Action (\textit{view2act})}\par
recover an action sequence that produces a target observation. &
\textbf{View2Location (\textit{view2loc})}\par
localize the agent (and optionally orientation) from a target observation under the map. \\
\bottomrule
\end{tabularx}

\caption{\textbf{Task suite comparison:} Route belief emphasizes egocentric, step-by-step path reasoning; Survey belief emphasizes allocentric mapping and novel view inference.}
\label{tab:route-survey-comparison}
\end{table}

\vspace{-3pt}
\subsection{Assessment Dimensions}
\label{sec:metrics}

We define assessment dimensions that align with the core \tos abilities: \textbf{construction} and \textbf{revision} are evaluated via exploration efficiency and belief quality, while \textbf{exploitation} is evaluated via task success.

\textbf{(D1) Belief Construction Efficiency.}~~\emph{Measures how efficiently the agent collapses spatial uncertainty during exploration.}
We quantify this using a normalized information gain metric, $\mathcal{E}$. Let $M$ be the number of possible positions for any object at the start of exploration (a uniform prior), and let $C_i$ be the number of positions for object $i$ that remain consistent with all observations gathered by the agent (calculated by AC-3 algorithm). The efficiency is calculated as  $\mathcal{E} = 1 - \frac{\sum_{i=1}^{N} \log_2 \max(1, C_i)}{N \log_2 M}$.
This score ranges from 0 (no information gained, $C_i=M$) to 1 (all objects perfectly localized, $C_i=1$).
Note that it can also be used to calculate the accumulated information gain at each step. 
Information gain is mainly used in text-based environments, since vision-based environments have direct access to scenes without such ambiguity. Therefore, for vision-based environments, we directly use node coverage to measure exploration efficiency.

\textbf{Belief Representation and Quality Assessment.}~~A core contribution of \tos is disentangling spatial memory from spatial inference. We structurally decompose the probed cognitive map into two components:

\begin{itemize}
    \item \textbf{(D2) The Cognitive Map (Observed):} \emph{Measures fidelity and coherent integration of observations over time.}
    We evaluate using two criteria: (1) Correctness, alignment with ground truth, computed as a composite of positional, directional, and facing accuracy; and (2) dynamic reasoning diagnostics, including \textbf{Perception} quality, \textbf{Self-tracking}, \textbf{Stability}, and \textbf{Local $\leftrightarrow$ Global Consistency}, reflecting internal coherence such as the absence of contradictions within the relational graph and between maps and relations.
    \item \textbf{(D3) The Uncertainty Map (Unobserved):} \emph{Measures how well the agent models plausible hypotheses about unobserved regions.}
    We assess \textbf{Uncertainty Modeling} by providing a candidate set of positions formed by randomly sampled points from both observed and unobserved areas, and measuring the agent’s ability to identify valid locations via $F_1$.
\end{itemize}
This separation lets us diagnose whether failures stem from \textit{misestimating} the observed world or from insufficient \textit{reasoning} about what remains unobserved.

\textbf{(D4) Belief Revision.}~~\emph{Measures the agent’s ability to revise its spatial belief under latent environment changes.}
We evaluate this using the \textbf{False Belief} task (\S\ref{sec:fb}), where objects are covertly manipulated (translated or rotated) following the initial exploration. The agent must re-explore to detect these discrepancies; we measure the accuracy of these identified changes (both object identity and transformation type) using the $F_1$ score. Furthermore, we introduce \textbf{Belief Inertia} to quantify whether belief revision remain biased toward obsolete priors.

\textbf{(D5) Belief Exploitation Success.}~~\emph{Measures task success when the agent must utilize its spatial belief.}
For tasks involving spatial relations (direction, persp.take, action2view), we score direction and distance separately, awarding 0.5 for each correct component. For tasks that output coordinates (view2loc, alloc.map), we compute a coordinate similarity score.

\subsection{Exploration Strategies}
\label{sec:passive-active}
To rigorously evaluate spatial cognition, we distinguish between two capabilities: the ability to acquire information (exploration) and the ability to synthesize it (reasoning). We present two evaluation settings: (i) \emph{Active Exploration}, where the agent must plan actions to reduce uncertainty, and (ii) \emph{Passive Comprehension}, where the agent reasons over standardized logs generated by scripted proxies.

\textbf{Uncertainty-Driven On-Policy Exploration. }
We conduct active evaluation to understand agent ability in \textbf{exploring the environment to gather necessary information in building spatial belief}. In this setting, the evaluated agent must plan and execute its own information-gathering policy. At each step, the agent selects an action based on its observation history and current objective, then receives new observations (text or image). Exploration continues until the agent issues an exploration \textit{termination} or reaches the step budget. Success requires balancing two goals: maximizing coverage of unknown relations while minimizing action cost. This setting directly reveals whether the agent can recognize what it does not yet know and actively reduce uncertainty through exploration. 

\textbf{Passive Exploration via Scripted Proxy Agents.} Evaluating \tos requires disentangling two intertwined factors: how well an agent explores, and how well it reasons about the observations gathered. An agent may fail either due to a suboptimal exploration policy (missing key evidence) or a deficiency in integrating observations into a coherent belief. To isolate the latter, we introduce \emph{proxy agents} as an exploration control.
In this setting, evaluated models are fed a fixed stream of observations generated by a proxy agent. By enforcing a standardized exploration path, we eliminate variance caused by exploration failures, allowing for a fair evaluation of core reasoning abilities across different architectures. 
We design two scripted proxies to provide standardized exploration logs. The \textbf{\textsc{Scout}} agent is used for \textit{visual environments}, who rotates at each location to guarantee all objects are observed. Leveraging visual cues like distance, these compact logs are sufficient for accurate belief construction. The \textbf{\textsc{Strategist}} agent is used for text environments, which follows a belief-driven edge-coverage policy and actively selects viewpoints to maximally reduce ambiguity in coarse symbolic observations. It is implemented with AC-3 constraint propagation to prune inconsistent hypotheses and ensure relations are uniquely determined. Implementation details for both agents appear in Appendix \P~\ref{par:proxy}. 

\section{Evaluation and Analysis}

We evaluate a set of state-of-the-art proprietary and open-source foundation models. They are evaluated on both passive and active settings described in \S~\ref{sec:passive-active}.
Unless otherwise specified for ablations, all experiments use three connected $6 \times 6$ rooms with $4$ objects in each (total $12$ objects). To enable a like-for-like comparison between the text and vision settings, we instantiate identical room layouts across modalities. 
We use $384\times384$ images in the vision setting. We generate 100 scenes and create three questions per task per scene, yielding $3\times 9 \times 100 = 2700$ questions per setting.
We mainly evaluate six foundation models: \textsc{GPT-5.2}~\citep{gpt-5.2}, \textsc{Gemini-3 pro}~\citep{gemini3pro}, \textsc{Claude-4.5-sonnet}~\citep{claude45sonnet}, \textsc{GLM-4.6V}~\citep{glm46v2025}, \textsc{Qwen3-VL}~\citep{bai2025qwen3vl} (235B-A22B-Thinking), and  \textsc{InternVL-3.5}~\citep{wang2025internvl3} (241B-A28B). For closed-source reasoning models \textsc{GPT-5.2}, \textsc{Gemini-3 Pro}, and \textsc{Claude-4.5-sonnet}, we set the temperature to $1$ and the maximum number of tokens to $32768$. For all other models, we set the temperature to 0. \textsc{InternVL-3.5} supports at most $10$ images, so we omit it for the vision-based world setting.

\begin{table*}[!htbp]
\centering
\small
\setlength{\tabcolsep}{6pt}
\setlength\extrarowheight{1pt}
\resizebox{0.95\textwidth}{!}{%
\begin{tabular}{l C{1.4\rescolwexp} *{8}{C{0.92\rescolwexp}} C{1.5\rescolwexp} @{} >{\columncolor{white}}m{0.35cm} !{\vrule width 0.5pt} @{} C{1.00cm}}
\multicolumn{1}{c}{} &  &
\roth{direction} & \roth{persp.take} & \roth{perc.dec.} & \roth{act2view} & \roth{view2act} &
\roth{alloc.map} & \roth{ment.rot} & \roth{loc2view} & \roth{view2loc} &
\multicolumn{1}{c}{} & \multicolumn{1}{c}{}\\

\multicolumn{1}{c}{} &
\multicolumn{1}{c}{} 
&
\multicolumn{1}{c}{\cellcolor{statbg}\textit{Static (S)}} &
\multicolumn{4}{c}{\cellcolor{dynbg}\textit{Dynamic (D)}} &
\multicolumn{1}{c}{\cellcolor{statbg}\textit{Static (S)}} &
\multicolumn{3}{c}{\cellcolor{dynbg}\textit{Dynamic (D)}} &
\multicolumn{1}{c}{} & \multicolumn{1}{c}{}\\

\textbf{Methods} &
\multicolumn{1}{c}{\cellcolor{surveybg}\textbf{Avg.step}}
&
\multicolumn{5}{c}{\cellcolor{routebg}\textbf{Route}} &
\multicolumn{4}{c}{\cellcolor{surveybg}\textbf{Survey}} &
\multicolumn{1}{c}{} & \textbf{Avg.}\\
\hline

\multicolumn{12}{c}{\textbf{Vision-based World}}\\
\hline
\rowcolor{secgray}\multicolumn{12}{l}{\textit{Proprietary Models}} \\
\textsc{GPT-5.2} & 17.2 & 40.0 & \textbf{36.7} & 56.2 & 43.8 & 40.3 & 43.4 & 59.7 & 56.9 & 37.8 & & 46.0 \\
\textsc{Gemini-3 Pro} & \textbf{13.6} & \textbf{56.3} & \textbf{36.7} & \textbf{68.2} & \textbf{47.2} & \textbf{54.0} & \textbf{63.5} & \textbf{73.0} & \textbf{65.4} & \textbf{52.2} & & \textbf{57.3} \\
\textsc{Claude-4.5 Sonnet}& 19.6 & 23.7 & 23.3 & 18.7 & 33.3 & 10.7 & 37.4 & 34.7 & 33.7 & 50.9 & & 29.6 \\
\hline
\rowcolor{secgray}\multicolumn{12}{l}{\textit{Open-source Models}} \\
\textsc{GLM-4.6V} & \textbf{15.0} & 15.8 & 18.5 & 3.3 & 14.0 & 0.7 & 18.9 & 8.0 & 18.5 & 31.8 & & 14.4 \\
\textsc{Qwen3-VL} & 16.3 & \textbf{16.8} & \textbf{23.3} & \textbf{13.4} & \textbf{24.8} & \textbf{5.7} & \textbf{25.8} & \textbf{16.3} & \textbf{21.5} & \textbf{43.7} & & \textbf{21.3} \\
\hline
\textsc{Human} & 9.8 & 94.5 & 100.0 & 100.0 & 100.0 & 93.4 & 93.4 & 100.0 & 100.0 & 86.7 & & 96.4 \\
\textsc{Human with Tool${}^\star$} & 11.1 & 100.0 & 100.0 & 100.0 & 100.0 & 97.8 & 100.0 & 100.0 & 100.0 & 93.4 & & 99.0 \\
\hline

\multicolumn{12}{c}{\textbf{Text-based World}}\\
\hline
\rowcolor{secgray}\multicolumn{12}{l}{\textit{Proprietary Models}} \\
\textsc{GPT-5.2} & \textbf{11.4} & 68.8 & 70.5 & 80.3 & 71.0 & 53.7 & 77.9 & 81.0 & 79.1 & 66.0 & & 72.0 \\
\textsc{Gemini-3 Pro} & 13.5 & \textbf{78.0} & \textbf{79.2} & \textbf{90.6} & \textbf{75.3} & \textbf{76.3} & \textbf{81.0} & \textbf{94.0} & \textbf{83.3} & \textbf{76.2} & & \textbf{81.5} \\
\textsc{Claude-4.5 Sonnet}& 18.7 & 65.3 & 65.3 & 79.0 & 62.7 & 51.7 & 68.8 & 76.3 & 57.0 & 67.0 & & 65.9 \\
\hline
\rowcolor{secgray}\multicolumn{12}{l}{\textit{Open-source Models}} \\
\textsc{GLM-4.6V} & 14.5 & 20.8 & 19.7 & 12.7 & 21.8 & 3.7 & 13.9 & 9.3 & 22.7 & 26.2 & & 16.8 \\
\textsc{InternVL-3.5} & 15.0 & 28.8 & 44.8 & 26.0 & \textbf{36.8} & 7.3 & 31.0 & 27.7 & 33.8 & 38.9 & & 30.6 \\
\textsc{Qwen3-VL} & \textbf{14.1} & \textbf{32.3} & \textbf{45.7} & \textbf{48.2} & 33.3 & \textbf{11.7} & \textbf{36.4} & \textbf{34.7} & \textbf{35.7} & \textbf{49.9} & & \textbf{36.8} \\
\hline
\textsc{Human}    & 10.8 & 87.8 & 82.1 & 100.0 & 85.5 & 86.8 & 66.6 & 100.0 & 95.6 & 75.8 & & 86.7 \\
\textsc{Human with Tool${}^\star$}    & 12.8 & 100.0 & 100.0 & 100.0 & 100.0 & 100.0 & 100.0 & 100.0 & 100.0 & 91.2 & & 99.0 \\
\hline

\end{tabular}
}
\captionsetup{type=table}
\caption{\textbf{Exploitation Performance ($\%$) of Belief Construction via Active Exploration.} Models autonomously plan actions and are evaluated on exploration cost, route-level reasoning, and survey-level reasoning across text- and vision-based environments. \textsc{Gemini-3 Pro} leads every task and all reasoning metrics, while \textsc{GPT-5.2} achieves the lowest exploration cost in text-world. Humans outperform in both settings, especially in vision. ${}^\star$Humans can use instruments such as protractors and compasses to infer object positions precisely.}

\label{tab:active}
\vspace{-10pt}
\end{table*}

\begin{table*}[!htbp]
\centering
\small
\setlength{\tabcolsep}{6pt}
\setlength\extrarowheight{1pt}
\resizebox{0.95\textwidth}{!}{%
\begin{tabular}{l *{8}{C{\rescolwexp}} C{1.7\rescolwexp} @{} >{\columncolor{white}}m{0.35cm} !{\vrule width 0.5pt} @{} C{1.00cm}}

\multicolumn{1}{c}{} &
\roth{direction} & \roth{persp.take} & \roth{perc.dec} & \roth{act2view} & \roth{view2act} &
\roth{alloc.map} & \roth{ment.rot} & \roth{loc2view} & \roth{view2loc} &
\multicolumn{1}{c}{} & \multicolumn{1}{c}{}\\

\multicolumn{1}{c}{} &
\multicolumn{1}{c}{\cellcolor{statbg}\textit{Static (S)}} &
\multicolumn{4}{c}{\cellcolor{dynbg}\textit{Dynamic (D)}} &
\multicolumn{1}{c}{\cellcolor{statbg}\textit{Static (S)}} &
\multicolumn{3}{c}{\cellcolor{dynbg}\textit{Dynamic (D)}} &
\multicolumn{1}{c}{} & \multicolumn{1}{c}{}\\

\textbf{Methods} &
\multicolumn{5}{c}{\cellcolor{routebg}\textbf{Route}} &
\multicolumn{4}{c}{\cellcolor{surveybg}\textbf{Survey}} &
\multicolumn{1}{c}{} & \textbf{Avg.}\\
\hline

\multicolumn{11}{c}{\textbf{Vision-based World}}\\
\hline
\rowcolor{secgray}\multicolumn{11}{l}{\textit{Proprietary Models}} \\

\textsc{GPT-5.2} &47.3 & 35.0 & \textbf{63.9} & \textbf{54.5} & 49.3 & 64.8 & 83.3 & 50.3 & \textbf{65.6} && 57.1 \\
\textsc{Gemini-3 Pro} & \textbf{63.8} & \textbf{36.3} & 57.5 & 49.0 & \textbf{58.0} & \textbf{67.2} & \textbf{85.3} & \textbf{70.4} & 57.0 && \textbf{60.5} \\
\textsc{Claude-4.5 Sonnet}& 47.3 & 33.5 & 37.7 & 40.8 & 15.7 & 54.8 & 58.3 & 44.7 & 54.8 && 43.1 \\
\hline
\rowcolor{secgray}\multicolumn{11}{l}{\textit{Open-source Models}} \\
\textsc{GLM-4.6V} & 11.5 & 24.5 & 4.7 & \textbf{19.0} & 2.7 & 22.9 & 11.7 & 20.0 & 33.6 && 16.7 \\
\textsc{Qwen3-VL} & \textbf{20.8} & \textbf{28.3} & \textbf{22.7} & 16.7 & \textbf{4.7} & \textbf{33.2} & \textbf{21.7} & \textbf{27.3} & \textbf{40.8} & & \textbf{24.9} \\
\hline

\multicolumn{11}{c}{\textbf{Text-based World}}\\
\hline
\rowcolor{secgray}\multicolumn{11}{l}{\textit{Proprietary Models}} \\

\textsc{GPT-5.2} & \textbf{84.5} & 88.2 & \textbf{97.0} & \textbf{89.0} & \textbf{76.0} & \textbf{96.3} & \textbf{98.3} & \textbf{94.8} & \textbf{89.2} && \textbf{90.4} \\
\textsc{Gemini-3 Pro} & 82.7 & \textbf{92.7} & \textbf{97.0} & 87.5 & 75.7 & 86.2 & 91.3 & 85.7 & 80.0 && 86.5 \\
\textsc{Claude-4.5 Sonnet}& 73.0 & 80.7 & 90.7 & 77.7 & 59.0 & 76.9 & 74.3 & 59.2 & 70.7 && 73.6 \\
\hline
\rowcolor{secgray}\multicolumn{11}{l}{\textit{Open-source Models}} \\

\textsc{GLM-4.6V} & 22.3 & 39.8 & 25.0 & 25.3 & 4.7 & 21.2 & 9.0 & 27.0 & 35.7 && 23.4 \\
\textsc{InternVL-3.5} & 36.7 & 67.8 & 42.7 & 41.2 & 8.7 & 37.3 & 19.3 & 38.7 & 43.8 & & 37.4 \\
\textsc{Qwen3-VL} & \textbf{40.8} & \textbf{69.3} & \textbf{56.5} & \textbf{50.0} & \textbf{17.7} & \textbf{42.8} & \textbf{40.3} & \textbf{42.5} & \textbf{54.6} & & \textbf{45.6} \\
\hline

\end{tabular}
}

\captionsetup{type=table}
\caption{\textbf{Exploitation Performance ($\%$) of Belief Construction via Passive Observations.} Models are evaluated as \emph{passive comprehension agents} on Route- and Survey-level reasoning using standardized observation logs from scripted proxy explorers, decoupling exploration from belief construction across text- and vision-based environments. \textsc{Gemini-3 Pro} leads most tasks in the vision-based world and achieves the best overall average, while \textsc{GPT-5.2} leads the text-based world and attains the best overall average.}
\label{tab:passive}
\vspace{-10pt}
\end{table*}

\textbf{Active Exploration Results.}
We evaluate models as active agents, where they must autonomously explore the environment to build their spatial belief and terminate the exploration process by their own. This setting tests the full \tos pipeline, requiring the agent to simultaneously plan an efficient information-gathering trajectory, integrate observations, and maintain a coherent cognitive map under uncertainty. 
The agent's performance is measured by its Exploration Efficiency as shown in \S~\ref{sec:metrics} and its final accuracy on the downstream spatial tasks. The agent has a maximum of $20$ exploration steps. Table~\ref{tab:active} presents the active performance of the models, providing a holistic view of their ability to translate curiosity into knowledge. Figure~\ref{fig:info_gain} illustrates information gain over the course of the exploration turns. \textsc{GPT-5.2} acquires substantial information early on, but its rate of gain slows in later turns, resulting in lower cumulative information gain than \textsc{Gemini-3 Pro} and \textsc{Claude-4.5 Sonnet}. Moreover, none of the models achieves full coverage relative to the proxy agent.
We benchmarked three human subjects across five text and five vision scenes. Humans consistently outperformed foundation models in both domains, particularly in vision. Intuitively, humans scored higher in vision than text as visual information is easier to process. With tools, they achieved near-perfect accuracy

\textbf{Passive Exploration Results.}
We evaluate models on trajectories generated by rule-based proxy agent to understand a model's core spatial reasoning ability regardless of its exploration strategy. 
The performance of various models in both text-based and vision-based environments is summarized in Table~\ref{tab:passive}. As evaluated, the results show a clear separation: \textsc{GPT-5.2} and \textsc{Gemini-3 Pro} lead by a wide margin over other systems, particularly open-source models. A substantial \textbf{modality gap} persists, with text performance far better than vision performance for all models. 

\begin{keytakeaways}{Modality Gap}
\begin{itemize}
  \item \textbf{Modality Gap Exists:} text significantly outperforms vision.
\end{itemize}
\end{keytakeaways}

Overall, active accuracies underperform the passive setting. Incomplete exploration leads to drops: Figure~\ref{fig:info_gain} shows that \textsc{GPT-5.2} gathers information quickly but often terminates prematurely, leaving uncertainty and lowering active scores relative to passive. Compared to the strategist proxy, which achieves full certainty, models remain less thorough. 
A second critical disparity is the efficiency gap. In the vision domain, the \textbf{\textsc{Scout}} proxy reaches target coverage in $\approx 9$ steps, whereas autonomous models expend significantly more actions with no performance benefit. This inefficiency is further highlighted in the text domain. While our primary text experiments utilize the \textbf{\textsc{Strategist}} proxy for maximum coverage, we additionally evaluated the \textbf{\textsc{Scout}} proxy in text world. The text-based \textbf{\textsc{Scout}} similarly averages $\approx 9$ steps. When following these concise trajectories, \textsc{GPT-5.2} and \textsc{Gemini-3 Pro} achieve accuracies of $83.9$ and $86.7$, respectively. These scores surpass their active exploration performance ($72.0, 81.5$ for \textsc{GPT-5.2} and \textsc{Gemini-3 Pro}, as in Table~\ref{tab:active}), demonstrating that models perform better when guided by a short, efficient proxy path than when exploring autonomously.

\begin{minipage}[b]{0.49\textwidth}
\centering
\tiny
\setlength{\tabcolsep}{3pt}
\setlength\extrarowheight{1pt}

\setlength{\rescolw}{0.05\linewidth} 

\resizebox{0.95\linewidth}{!}{%
\begin{tabular}{l *{3}{C{\rescolw}} | *{3}{C{\rescolw}}}
\hline
\multicolumn{7}{c}{\textbf{Text-based World}}\\
\hline
\textbf{Methods} &
\multicolumn{3}{c|}{\textbf{2-room}} &
\multicolumn{3}{c}{\textbf{4-room}} \\
\hline
& \textit{pass.} & \textit{act.} & \textit{steps} &
  \textit{pass.} & \textit{act.} & \textit{steps} \\
\hline
\textsc{GPT-5.2}      & \textbf{92.3} & 77.8 & \textbf{6.2} & \textbf{86.5} & 66.0 & \textbf{16.4} \\
\textsc{Gemini-3 Pro} & 86.7 & \textbf{80.6} & \textbf{6.2} & 81.2 & \textbf{77.7} & 19.7 \\
\hline
\multicolumn{7}{c}{\textbf{Vision-based World}}\\
\hline
\textbf{Methods} &
\multicolumn{3}{c|}{\textbf{2-room}} &
\multicolumn{3}{c}{\textbf{4-room}} \\
\hline
& \textit{pass.} & \textit{act.} & \textit{steps} &
  \textit{pass.} & \textit{act.} & \textit{steps} \\
\hline
\textsc{GPT-5.2}      & \textbf{59.3} & 51.5 & 10.8 & 52.6 & 40.3 & 23.2 \\
\textsc{Gemini-3 Pro} & 58.3 & \textbf{57.8} & \textbf{6.6} & \textbf{56.2} & \textbf{51.5} & \textbf{19.7} \\
\hline
\end{tabular}
}

\captionof{table}{\textbf{Exploitation Performance ($\%$) for Multi-Room Settings} (2-room and 4-room). %
\textit{pass.} for passive avg acc, \textit{act.} for active avg acc, \textit{steps} for average steps.}
\label{tab:multi-rooms}
\vspace{-6pt}
\end{minipage}
\hfill
\begin{minipage}[b]{0.44\textwidth}
\centering
\includegraphics[width=\linewidth]{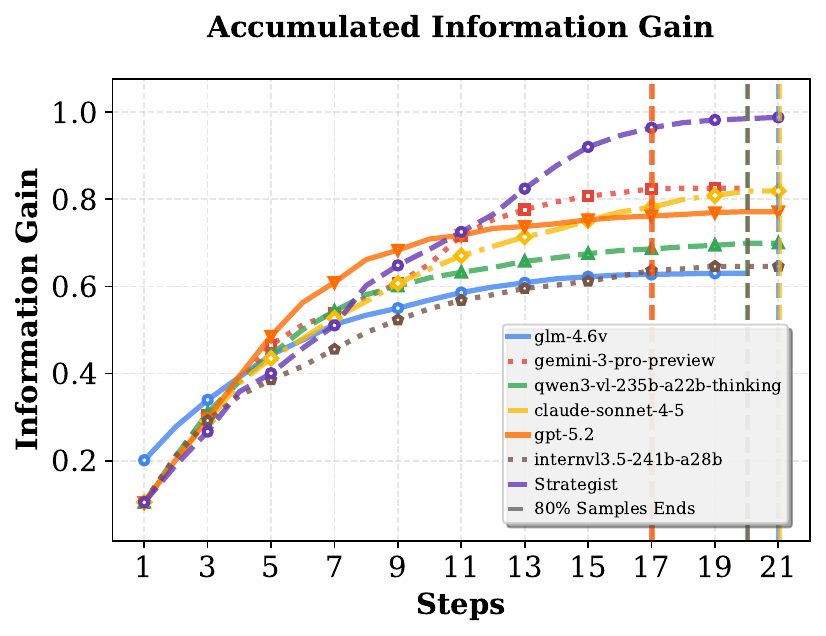}
\captionof{figure}{\textbf{Accumulated information gain} over exploration steps in the text world.}
\label{fig:info_gain}
\vspace{-6pt}
\end{minipage}

\textbf{Different Room Settings.}~
For the two best-performing models, \textsc{GPT-5.2} and \textsc{Gemini-3 Pro}, we further evaluate reasoning and exploration under different room configurations: a four-room setting and two three-room settings. In the four-room setting, the main room connects to the other three rooms.
Table~\ref{tab:multi-rooms} reports results across different room settings. As the number of rooms increases, exploration cost rises accordingly. For both \textsc{GPT-5.2} and \textsc{Gemini-3 Pro}, performance declines as the room number increases, and the active–passive performance gap widens with room number. Moreover, \textsc{Gemini-3 Pro} requires nearly the same number of exploration steps in the text-only and vision-based environments.
Detailed results are in Appendix \P~\ref{par:additional_results}.

\begin{keytakeaways}{Active Exploration as the Bottleneck}
\begin{itemize}\setlength{\itemsep}{1pt}
\item \textbf{Performance and Efficiency Deficit:} Active agents score lower than reasoning on rule based program histories, and explore less efficiently than the program.
\item \textbf{Incomplete Coverage:} Active agent fails to achieve complete information coverage.
\item \textbf{Complexity-Widened Gap:} The active versus passive difference grows with environment scale; \textsc{Gemini-3 Pro} degrades least.
\end{itemize}
\end{keytakeaways}

\textbf{Exploration Pattern} 
Manual inspection of agent exploration histories reveals distinct behavioral patterns. For \textsc{GPT-5.2}, the active-passive performance gap stems from unsystematic exploration. Specifically, the agent tends to prioritize any newly discovered door, immediately jumping to inspect it and often leaving the current room partially unexplored. This is compounded by object omission and path redundancy. In contrast, \textsc{Gemini-3 Pro} adopts a more methodical ``rotate-and-scan" strategy, scanning its surroundings before transitioning to new rooms, which is a behavior mirroring the \textbf{\textsc{Scout}} proxy agent. Further examples are provided in Appendix \P~\ref{par:exp_pattern}.
\section{How do Foundation Models Manage Internal Spatial Belief?}

\begin{figure}[t]
  \centering
    \centering
    \includegraphics[width=0.9\linewidth]{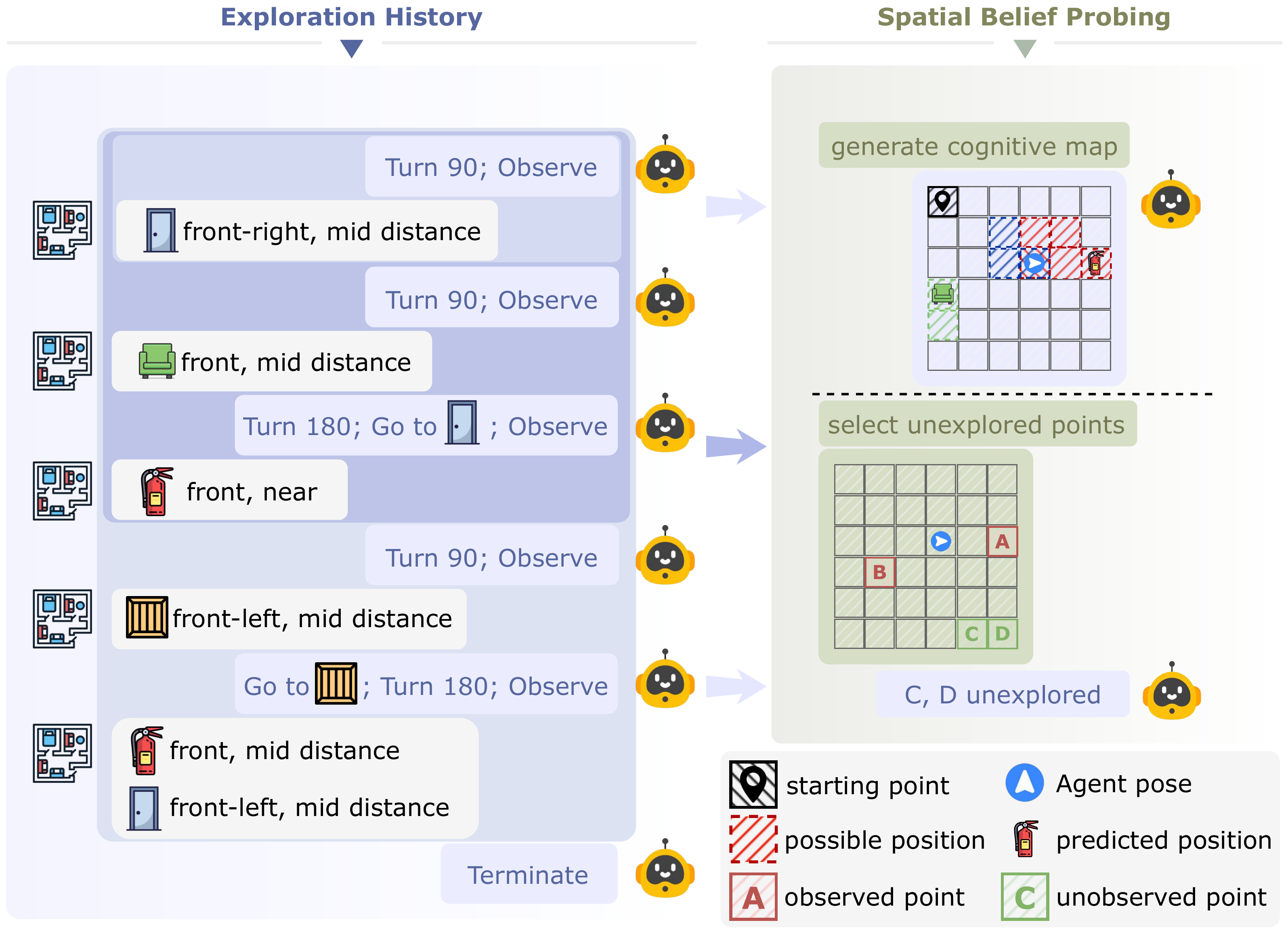}
    \caption{
    \textbf{Internal Spatial Belief Probing.} At each step, the agent executes an action, receives an observation, and updates its spatial belief. We probe this belief by prompting the agent to (i) output a JSON-structured cognitive map of all observed objects and (ii) select the next unexplored position from a top-down view given a set of labeled candidate points. For clarity, the figure shows the probing process for a single step.
    }
    \label{fig:spatial_belief}
\end{figure}

In this section, we use the \tos belief-probing mechanism (as proposed in \S\ref{sec:tos4llm}) to diagnose how MLLMs manage internal spatial beliefs and move beyond treating the agent as a black box. Figure~\ref{fig:spatial_belief} shows the example of how we probe the belief of agent at each exploration step

\subsection{Cognitive Map Probing}
\label{sec:cogmap_probing}
Instead of treating the spatial belief as a black box, we probe the agent’s internal state to distinguish verifying known facts from hypothesizing about the unknown. The agent externalizes its belief via a structured JSON containing a \textbf{Cognitive Map}, which records objects currently or previously observed within the field of view.

\textbf{Representation.} For consolidated map, the agent presents its belief as a single, allocentric cognitive map serialized in structured JSON. The map maintains (i) a \emph{global} layout anchored to the agent’s initial pose, and (ii) a \emph{local} snapshot that records only the currently visible objects with the current pose as origin to diagnose immediate perceptual errors.  

\textbf{Metrics.} We evaluate consolidated map using three complementary metrics. 
\emph{Positional accuracy (pos.acc)} is the Euclidean similarity between predicted and true object coordinates: 
$(K/N) \cdot e^{-\mathrm{RMSE} / L}$, where $\mathrm{RMSE}$ is the root mean squared error between predicted and ground-truth object positions, $L$ is the RMS $\ell_2$-norm of the positions of all objects in the scene, and $K/N$ is the coverage (the ratio of the number of predicted objects K to the number of ground-truth objects N).
\emph{Directional accuracy (dir.acc)} is the accuracy of directional relationship between each pair of objects.
\emph{Facing accuracy (facing.acc)} is the fraction of objects whose predicted facing matches the ground truth. 

Using \emph{global} and \emph{local} belief representations, we compute a set of diagnostic scores at each turn $t$ (all per-turn except \textbf{Correctness}, which is computed only at the final turn after termination). Unless noted, scores are averaged over turns and scenes:

\begin{itemize}
  \item \textbf{Correctness (final)}: \emph{Measures the accuracy of the agent’s terminal global spatial belief.}
  At the last turn, we evaluate the predicted global map and report a composite score given by the (equally weighted) mean of the three metrics defined above, with weights $1/3$ each. We compute \emph{dir.acc} only for correctness, since the global cognitive map prioritizes consistent pairwise spatial relations.

  \item \textbf{Perception}: \emph{Measures how accurately the agent interprets newly observed local structure.}
  We compare the predicted local map to the ground-truth local map for the current field of view (FOV), counting only objects that appear in the FOV for the first time.

  \item \textbf{Self-tracking}: \emph{Measures how well the model estimates its own pose over time.}
  We infer the agent's pose from the predicted global map and compare it against the ground-truth agent state.

  \item \textbf{Local $\leftrightarrow$ Global consistency}: \emph{Measures whether new local evidence is incorporated into the global belief coherently.}
  Within the same turn, we compare local and global predictions to verify that newly perceived structure is integrated without contradictions.

  \item \textbf{Stability}: \emph{Measures whether beliefs about previously observed objects remain non-degrading over time.}
  For each previously observed object, at every subsequent turn we check that its predicted state does not worsen; the per-check score is $1$ if the prediction is no worse than in the previous turn.
\end{itemize}

Results in Table~\ref{tab:belief-probe} indicate a substantial modality gap between vision and text: performance drops markedly in the vision setting across all metrics, not just belief \textbf{Correctness}.
\textbf{Self-tracking} does not appear to be a primary bottleneck, models can often maintain an accurate belief about their own pose.
\textbf{Perception} remains a key limitation for state-of-the-art models in visual world settings. In particular, recognizing an object’s facing direction is especially challenging: agents frequently fail to infer orientation and achieve near-chance (or worse) facing \textbf{Correctness}. This weakness is consistent with Table~\ref{tab:active}, where agents perform poorly on perspective-taking tasks (about $36\%$ accuracy).
\textbf{Stability \& Decay.}
Crucially, the metric reveals that spatial beliefs are highly brittle not just for orientation, but also for position.
While \textbf{Perception} scores indicate that models can capture local spatial details with reasonable accuracy, this initial fidelity fails to translate into final map \textbf{Correctness}.
This performance gap highlights a critical failure in state maintenance: even when objects are correctly perceived initially, the agent frequently overwrites these verified facts with incorrect predictions in subsequent turns.
Thus, the low final Correctness stems not solely from perceptual errors, but from the cumulative effect of unstable belief updates, where valid spatial memories degrade over the course of the episode.

\begin{table*}[!tbp]
\centering
\small
\setlength{\tabcolsep}{3.5pt}
\renewcommand{\arraystretch}{1.25}

\colorlet{perbg}{violet!15}

\setlength{\aboverulesep}{0pt}
\setlength{\belowrulesep}{0pt}

\resizebox{0.9\textwidth}{!}{%
\begin{tabular}{
>{\raggedright\arraybackslash}m{0.140\textwidth} % Methods
>{\centering\arraybackslash}m{0.050\textwidth} % Correctness 1 (Ori)
>{\centering\arraybackslash}m{0.050\textwidth} % Correctness New (Pos) <--- NEW COLUMN
>{\centering\arraybackslash}m{0.050\textwidth} % Correctness 2 (Overall)
>{\centering\arraybackslash}m{0.045\textwidth} % Perception (New) 1
>{\centering\arraybackslash}m{0.045\textwidth} % Perception (New) 2
% Perception (All) Removed
>{\centering\arraybackslash}m{0.055\textwidth} % Local-Global 1
>{\centering\arraybackslash}m{0.055\textwidth} % Local-Global 2
>{\centering\arraybackslash}m{0.050\textwidth} % Stability 1
>{\centering\arraybackslash}m{0.050\textwidth} % Stability 2
>{\centering\arraybackslash}m{0.060\textwidth} % Self-tracking 1 (New Split: Ori)
>{\centering\arraybackslash}m{0.060\textwidth} % Self-tracking 2 (New Split: Pos)
>{\centering\arraybackslash}m{0.110\textwidth}  % Uncertainty
}

% === Header Row 1: Sub-categories ===
\multicolumn{1}{c}{} &
\roth{\footnotesize ori.} & \roth{\footnotesize pos.} & \roth{\footnotesize overall} &
\roth{\footnotesize ori.} & \roth{\footnotesize pos.} & 
\roth{\footnotesize ori.} & \roth{\footnotesize pos.} & 
\roth{\footnotesize ori.} & \roth{\footnotesize pos.} &
\roth{\footnotesize ori.} & \roth{\footnotesize pos.} &
\multicolumn{1}{c}{} \\
\multicolumn{1}{c}{\bfseries\footnotesize Methods} &
\multicolumn{3}{c}{\cellcolor{black!5}\bfseries\footnotesize Correctness ($\%$)} & 
\multicolumn{2}{c}{\cellcolor{black!7}\bfseries\footnotesize Perception ($\%$)} & 
\multicolumn{2}{c}{\cellcolor{black!5}\parbox{0.11\textwidth}{\centering\bfseries\footnotesize Local$\leftrightarrow$\newline Global ($\%$)}} & 
\multicolumn{2}{c}{\cellcolor{black!7}\bfseries\footnotesize Stability ($\%$)} &
\multicolumn{2}{c}{\cellcolor{black!5}\parbox{0.12\textwidth}{\centering\bfseries\footnotesize Self-\newline tracking ($\%$)}} & 
\bfseries\footnotesize Uncertainty ($\%$)  \\ 
\hline

\multicolumn{13}{c}{\bfseries Vision-based World}\\ 
\hline
\mbox{\textsc{GPT-5.2}}
& 20.2 & 42.0 & \multicolumn{1}{>{\centering\arraybackslash}m{0.050\textwidth}|}{32.2} 
& 33.5 & \textbf{72.4} 
& \textbf{57.9} & 58.7 
& \textbf{65.4} & 56.4
& 93.3 & 64.7
& \multicolumn{1}{|>{\centering\arraybackslash}m{0.100\textwidth}}{53.7} \\

\mbox{\textsc{Gemini-3 Pro}}
& \textbf{32.2} & \textbf{62.5} & \multicolumn{1}{>{\centering\arraybackslash}m{0.050\textwidth}|}{\textbf{52.1}}
& \textbf{43.8} & 68.5 
& 52.9 & \textbf{68.3}
& 61.8 & \textbf{62.0}
& \textbf{98.8} & \textbf{73.9}
& \multicolumn{1}{|>{\centering\arraybackslash}m{0.100\textwidth}}{\textbf{70.2}} \\
\hline

\multicolumn{13}{c}{\bfseries Text-based World}\\
\hline
\mbox{\textsc{GPT-5.2}}
& 91.0 & 75.1 & \multicolumn{1}{>{\centering\arraybackslash}m{0.050\textwidth}|}{80.0} 
& 100 & 86.8 
& \textbf{96.4} & \textbf{86.0}
& \textbf{96.7} & 67.6
& 98.0 & \textbf{86.7}
& \multicolumn{1}{|>{\centering\arraybackslash}m{0.100\textwidth}}{64.5} \\

\mbox{\textsc{Gemini-3 Pro}}
& \textbf{92.5} & \textbf{75.5} & \multicolumn{1}{>{\centering\arraybackslash}m{0.050\textwidth}|}{\textbf{81.4}} 
& 99.9 & \textbf{88.2} 
& 91.6 & 84.8
& 90.8 & 67.7
& \textbf{99.9} & 85.2
& \multicolumn{1}{|>{\centering\arraybackslash}m{0.100\textwidth}}{\textbf{79.2}} \\
\hline
\end{tabular}}
\caption{\textbf{Spatial Belief Quality via Cognitive Map Probing.}
We measure final map correctness and turn-level perception, local global consistency, stability, self-tracking, and uncertainty in text- vs. vision-worlds. \textit{ori.} for orientation and \textit{pos.} for position.
Across models, vision lags text on all metrics, with the largest drop on orientation and stability.}
\label{tab:belief-probe}
\vspace{-10pt}
\end{table*}

\begin{keytakeaways}{Cognitive Map Failures (Orientation, Stability, and Belief Drift)}
\begin{itemize}
    \setlength{\itemsep}{1pt}
    \item \textbf{Orientation Gap:} Vision perception is a bottleneck, especially for object orientation.
    \item \textbf{Unstable Map:} Beliefs about previously observed objects degrades over time.
    \item \textbf{Belief Drift:} New updates corrupt earlier correct perceptions, lowering final correctness.
\end{itemize}
\end{keytakeaways}

\textbf{Cognitive Map Validation \& Correlation.} To validate the utility of the probed cognitive map and investigate whether it faithfully reflects the agent's reasoning process, we first conducted two ablation studies:
\begin{itemize}[leftmargin=*]
    \item \textbf{Sufficiency Test (Oracle Map):} We conditioned the model on the ground-truth cognitive map before generating answers for evaluation. Performance rose to near-perfect levels ($\approx95\%$ for both models in both worlds). This confirms that our cognitive map representation captures \emph{all} necessary information for the tasks; performance bottlenecks stem from the agent's inability to accurately \emph{construct} the map, not the representation format itself.
    \item \textbf{Alignment Test (Explicit Reasoning):} We prompted the model to explicitly generate the cognitive map before answering the evaluation questions. This resulted in a slight performance degradation compared to direct answering.
\end{itemize}
These results reveal an \textbf{externalization gap}: the model's latent internal spatial belief is richer or more accurate than the discretized JSON output it produces. \textbf{While it is a lossy compression of the agent's true internal state, the explicit map remains a strong diagnostic signal.}
We support this claim by computing the Pearson correlation between the agent’s cognitive map \textbf{Correctness} and downstream task performance. To ensure a robust correlation, we calculate the average performance across five independent cognitive map runs for each sample.
As shown in Table~\ref{tab:correlation}, belief correctness is consistently and positively correlated with downstream success in both modalities, with all correlations significant ($p<.001$). The association is stronger in vision ($r{=}0.570/0.645$) than in text ($r{=}0.418/0.466$). The stronger vision correlation suggests that perception-driven mapping errors and unstable belief updates more directly translate into task failures. Thus, we establish map probing as a \emph{validated diagnostic proxy} for failure analysis. While acknowledging that correlation does not imply causality, we treat the explicit map as a robust, albeit conservative, signal for diagnosing reasoning breakdowns rather than definitive evidence.

\begin{keytakeaways}{Maps as a Diagnostic Proxy}
\begin{itemize}
    \setlength{\itemsep}{1pt}
    \item \textbf{Lossy but Diagnostic:} Though a lossy compression, map correctness correlates significantly with downstream success, making it a strong diagnostic signal.
\end{itemize}
\end{keytakeaways}

% \begin{figure}[t]
%   \centering

%   \begin{subfigure}[t]{0.48\linewidth}
%     \centering
%     \footnotesize
%     \setlength{\tabcolsep}{6pt}
%     \renewcommand{\arraystretch}{1.1}
%     \begin{tabular}{l c c}
%       \hline
%       \textbf{Methods} & \textbf{Text} & \textbf{Vision} \\
%       \hline
%       \textsc{GPT-5.2}      & 41.8 & 57.0 \\
%       \textsc{Gemini-3 Pro} & 46.6 & 64.5 \\
%       \hline
%     \end{tabular}
%     \caption{Pearson correlation ($r$) between spatial-belief correctness and downstream evaluation performance. All correlations are significant ($p<.001$).}
%     \label{fig:correlation}
%   \end{subfigure}\hfill%
%   \begin{subfigure}[t]{0.48\linewidth}
%     \centering
%     \includegraphics[width=\linewidth]{Arxiv/figures/info_gain_cogmap.pdf}
%     \caption{Accumulated information gain and cognitive map correctness over steps for \textsc{GPT-5.2} and \textsc{Gemini-3 Pro}.}
%     \label{fig:info_gain_cogmap}
%   \end{subfigure}

% \end{figure}

\subsection{Uncertainty Map Probing}

\begin{wraptable}[10]{r}{0.48\textwidth}
  \centering
  \footnotesize
  \setlength{\tabcolsep}{8pt}
  \renewcommand{\arraystretch}{1.1}
  \begin{tabular}{l c c}
    \hline
    \textbf{Methods} & \textbf{Text ($\%$)} & \textbf{Vision ($\%$)} \\
    \hline
    \textsc{GPT-5.2}      & 41.8 & 57.0 \\
    \textsc{Gemini-3 Pro} & 46.6 & 64.5 \\
    \hline
  \end{tabular}
  \caption{\textbf{Pearson correlation} ($r$) between spatial-belief correctness and downstream evaluation performance. All correlations are significant ($p<.001$).}
  \label{tab:correlation}
\end{wraptable}
To probe an agent’s ability to model uncertainty, we provide it with a top-down view of the scene in which all objects are removed, and we overlay a set of candidate points. These points are sampled randomly and include both previously observed and unobserved locations. The agent’s task is to identify which candidate points remain unobserved, thereby revealing its belief over unseen regions.
\begin{wrapfigure}{r}{0.45\textwidth}
  \centering
  \includegraphics[width=\linewidth]{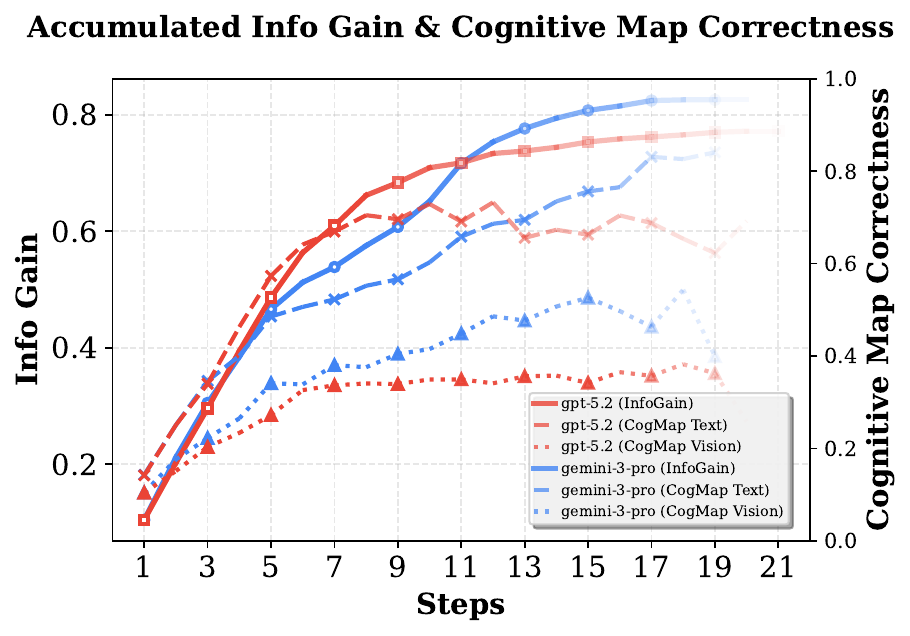}
  \caption{Accumulated Information Gain and Cognitive Map Correctness over steps.}
  \label{fig:info_gain_cogmap}
\end{wrapfigure}

\textbf{Representation.} The agent receives an empty top down map that shows only the candidate points and its current position, with no objects present. The agent must select the points that have not yet been observed. 
In the text based world, the top down map is represented as an $N \times M$ symbolic grid, where different symbols denote the agent, gates, and candidate points. In the vision based world, all objects are removed and the agent instead receives a top down image of the environment, check examples in Appendix \P~\ref{sec:app:prompts}. We use $F_1$ to evaluate selected points.

We report \textbf{Uncertainty} scores in Table~\ref{tab:belief-probe}. \textsc{Gemini-3 Pro} models uncertainty better than \textsc{GPT-5.2} in both text- and vision-based settings. These results help explain the information gain and cognitive map trends in Figure~\ref{fig:info_gain_cogmap}. \textsc{GPT-5.2} achieves higher initial information gain (i.e., it ramps up faster), likely because it quickly commits to an explore-the-doors strategy. However, it generalizes poorly to unobserved regions, reflected by the subsequent plateau in Figure~\ref{fig:info_gain_cogmap}: additional steps yield little marginal gain. In contrast, although \textsc{Gemini-3 Pro} improves more slowly at the beginning, its cognitive map accuracy continues to increase with exploration, suggesting it keeps collecting useful evidence and progressively resolving uncertainty.

\subsection{Belief Revision Task}
\label{sec:fb}
Spatial intelligence requires not only mapping static environments but also maintaining beliefs under non-stationarity. Inspired by \textit{false belief} protocols in developmental psychology \citep{wimmer1983belief, baron1985does} and \textit{spatial belief revision}~\citep{knauff2013spatial}, we introduce a dynamic perturbation task to probe the agent's ability to discard obsolete priors and reintegrate new evidence.

\textbf{Task Protocol.} Following the initial exploration phase, we introduce a discrete environmental shift: a subset of $k=4$ objects are stochastically relocated or reoriented. The agent, retaining its memory (exploration history), must actively re-explore the environment to identify the state changes. This requires the agent to detect conflicts between its internal belief state and new sensory observations.

\textbf{Metrics.} We evaluate performance along four complementary axes:

\begin{itemize}
    \item \textbf{Identification Accuracy ($F_1$):} \emph{How precisely the agent pinpoints which objects changed.}
    We compute the $F_1$ score for detecting the subset of objects whose position or orientation shifted.

    \item \textbf{Average Steps:} \emph{How efficiently the agent revises its beliefs to completion.}
    We report \textit{Total Steps} needed to identify all changes, and \textit{Redundancy Steps}, defined as the number of steps taken after the last changed object has been observed. Ideally, Redundancy $\to 0$, indicating the agent recognizes when updating is complete.

    \item \textbf{Belief Correctness:} \emph{How accurate the updated beliefs are on the changed subset.}
    We compute correctness as in \S\ref{sec:cogmap_probing}, but restrict evaluation to changed objects to isolate the fidelity of re-exploration.

    \item \textbf{Belief Inertia:} \emph{Whether updating remains systematically biased toward obsolete priors.}
    To quantify attraction back to pre-shift beliefs, we test whether the residual error of the updated belief aligns with the direction of the \emph{old} belief. For each shifted object $i$, let $\mathbf{b}_i^{old}$ denote the pre-shift belief, $\mathbf{b}_i^{new}$ the post-revision belief, and $\mathbf{g}_i^{new}$ the post-shift ground truth. Define the \textit{prior-offset} and \textit{post-revision error} vectors:
    $
    \mathbf{v}_i = \mathbf{b}_i^{old} - \mathbf{g}_i^{new}, 
    \mathbf{e}_i = \mathbf{b}_i^{new} - \mathbf{g}_i^{new}.
    $
    We define positional inertia as
    \[
    s_i^{pos}
    = 
    \underbrace{\frac{\mathbf{e}_i^\top \mathbf{v}_i}{\|\mathbf{e}_i\|\,\|\mathbf{v}_i\| + \epsilon}}_{\text{Directional alignment }(\cos\theta_i)}
    \cdot
    \underbrace{\exp\!\left(-\frac{\|\mathbf{b}_i^{new}-\mathbf{b}_i^{old}\|^2}{2\sigma^2}\right)}_{\text{Proximity weight }(w_i)} .
    \]
    Here $\cos\theta_i$ is large when the remaining error after updating still points toward the obsolete location, while $w_i$ downweights such alignment when the belief has moved far from $\mathbf{b}_i^{old}$. We set $\sigma$ to a dynamic noise scale: the RMS localization error on the first re-observed \emph{unchanged} objects during re-exploration; $\epsilon$ ensures numerical stability. Under unbiased updating, $\mathbb{E}[s_i^{pos}] \approx 0$, whereas $s_i^{pos} > 0$ indicates systematic pull toward the obsolete prior.
    For orientation shifts, we measure inertia via 
    $
    s_i^{ori}=\mathds{1}\!\left(\phi_i^{new}=\phi_i^{old}\right),
    $
    where $\phi$ denotes the predicted orientation. It flags failures to overwrite the obsolete facing direction.
\end{itemize}

Table~\ref{tab:fb-belief-probe} corroborates the modality gap observed in previous sections: vision-based agents significantly underperform their text-based counterparts. This performance drop is characterized by increased exploration redundancy and lower accuracy in identifying changed objects. 
Notably, while belief inertia persists across both modalities, it is markedly more severe in vision-based agents, particularly regarding object orientation. Vision models frequently fail to overwrite their initial spatial memory, persisting with obsolete facing estimates despite new visual evidence. This also suggests that fine-grained orientation estimation remains a critical bottleneck for visual spatial reasoning.

\begin{table*}[!tbp]
\centering
\small
\setlength{\tabcolsep}{3.5pt}
\renewcommand{\arraystretch}{1.25}

\colorlet{perbg}{violet!15}

\setlength{\aboverulesep}{0pt}
\setlength{\belowrulesep}{0pt}

\resizebox{0.95\textwidth}{!}{%
\begin{tabular}{
>{\raggedright\arraybackslash}m{0.140\textwidth}
>{\centering\arraybackslash}m{0.080\textwidth}
>{\centering\arraybackslash}m{0.080\textwidth}
>{\centering\arraybackslash}m{0.080\textwidth}
>{\centering\arraybackslash}m{0.080\textwidth}
>{\centering\arraybackslash}m{0.09\textwidth}
>{\centering\arraybackslash}m{0.09\textwidth}
>{\centering\arraybackslash}m{0.080\textwidth}
>{\centering\arraybackslash}m{0.080\textwidth}
}

\multicolumn{1}{c}{} &
\roth{\footnotesize all} & \roth{\footnotesize red.} &
\roth{\footnotesize ori.} & \roth{\footnotesize pos.} & 
\roth{\footnotesize ori.} & \roth{\footnotesize pos.} &
\roth{\footnotesize ori.} & \roth{\footnotesize pos.} \\

\multicolumn{1}{c}{\bfseries\footnotesize Methods} &
\multicolumn{2}{c}{\cellcolor{black!5}\bfseries\footnotesize Avg. Steps $\downarrow$} &
\multicolumn{2}{c}{\cellcolor{black!7}\bfseries\footnotesize Identification ($\%$) $\uparrow$ } & 
\multicolumn{2}{c}{\cellcolor{black!5}\parbox{0.18\textwidth}{\centering\bfseries\footnotesize Belief \newline Correctness  ($\%$) $\uparrow$}} &
\multicolumn{2}{c}{\cellcolor{black!7}\parbox{0.15\textwidth}{\centering\bfseries\footnotesize Belief \newline Inertia ($\%$) $\downarrow$}} \\ 
\hline

\multicolumn{9}{c}{\bfseries Text-based World}\\
\hline
\mbox{\textsc{GPT-5.2}}
& \textbf{6.92} & \multicolumn{1}{>{\centering\arraybackslash}m{0.080\textwidth}|}{0.55}
& 97.9 & \multicolumn{1}{>{\centering\arraybackslash}m{0.080\textwidth}|}{98.4}
& 89.5 & \multicolumn{1}{>{\centering\arraybackslash}m{0.080\textwidth}|}{69.7}
& \textbf{5.5} & 12.5 \\

\mbox{\textsc{Gemini-3 Pro}}
& 7.79 & \multicolumn{1}{>{\centering\arraybackslash}m{0.080\textwidth}|}{\textbf{0.18}} 
& \textbf{98.7} & \multicolumn{1}{>{\centering\arraybackslash}m{0.080\textwidth}|}{\textbf{98.8}}
& \textbf{91.8} & \multicolumn{1}{>{\centering\arraybackslash}m{0.080\textwidth}|}{\textbf{72.9}}
& 7.9 & \textbf{5.7} \\ 
\hline

\multicolumn{9}{c}{\bfseries Vision-based World}\\
\hline
\mbox{\textsc{GPT-5.2}}
& 13.06 & \multicolumn{1}{>{\centering\arraybackslash}m{0.080\textwidth}|}{6.20}
& 14.3 & \multicolumn{1}{>{\centering\arraybackslash}m{0.080\textwidth}|}{68.0}
& 16.7 & \multicolumn{1}{>{\centering\arraybackslash}m{0.080\textwidth}|}{42.9}
& 68.9 & 34.7 \\ 

\mbox{\textsc{Gemini-3 Pro}}
& \textbf{10.29} & \multicolumn{1}{>{\centering\arraybackslash}m{0.080\textwidth}|}{\textbf{3.23}}
& \textbf{23.9} & \multicolumn{1}{>{\centering\arraybackslash}m{0.080\textwidth}|}{\textbf{82.5}}
& \textbf{30.3} & \multicolumn{1}{>{\centering\arraybackslash}m{0.080\textwidth}|}{\textbf{63.1}}
& \textbf{51.1} & \textbf{14.4} \\
\hline
\end{tabular}}
\caption{\textbf{Belief updating under environmental shifts.}
After relocating/reorienting $k{=}4$ objects, we evaluate change identification, re-exploration cost (including redundancy (red.)), and belief correctness/update in text- vs. vision-worlds.
Vision agents require more redundant steps and show severe orientation inertia, failing to overwrite obsolete facing beliefs despite new evidence.}
\label{tab:fb-belief-probe}
\vspace{-10pt}
\end{table*}

\begin{keytakeaways}{Vision Deficiencies \& Belief Inertia}
\begin{itemize}
    \setlength{\itemsep}{1pt}
    \item \textbf{Vision-based Revision Failures:} Vision agents suffer from excessive exploration redundancy and poor accuracy in identifying object shifts.
    \item \textbf{Belief Inertia:} Agents, especially vision-based ones, persist in obsolete spatial coordinates despite new observations.
\end{itemize}
\end{keytakeaways}

\section{Related Work}
\textbf{Passive Spatial Reasoning.}
Early paradigms treat spatial reasoning as static inference: given a textual description, agents answer relational queries~\citep{weston2015towards, shi2022stepgame, mirzaee2021spartqa, li2024reframing}. Other benchmarks probe understanding from a single image, asking for relative directions, topological relations, or metric attributes~\citep{ma20243dsrbench, deng2025internspatial, cheng2024spatialrgpt, chen2024spatialvlm, liao2024reasoning, kamath2023s}. Multi-view and video benchmarks raise difficulty by requiring cross-view integration, egocentric–allocentric conversion, and temporal consistency~\citep{yang2025mmsi, xu2025multi, wu2025spatialscore, yeh2025seeing, gholami2025spatial, zhou2025vlm4d}. Recent works explicitly adopt cognitive maps: VSI-Bench~\citep{yang2025thinking} shows map formation improves video QA, and MindCube~\citep{yin2025spatial} demonstrates that predicting layouts boosts multi-view reasoning. While informative, these benchmarks remain disembodied, as agents reason only over pre-collected trajectories.

\textbf{Active Exploration for Spatial Understanding.}
Research has also examined agents that actively explore, but their exploration is usually tied to task-specific goals rather than building a general spatial belief. Embodied question answering benchmarks evaluate agents by whether they can gather evidence to answer questions~\citep{das2018embodied,gordon2018iqa,majumdar2024openeqa,ginting2025entermindpalacereasoning,ren2024explore}. Instruction-following settings extend household tasks to long horizons and realistic scenes, often with dialog or language grounding~\citep{shridhar2020alfred,kim2024realfred,shridhar2020alfworld,puig2018virtualhome,padmakumar2022teach,gao2022dialfred}. Navigation benchmarks stress path execution and generalization across diverse environments~\citep{anderson2018vision,jain2019stay,ku2020rxr,krantz2020beyond,nguyen2019help,wang2024divscene,zhao2025cityeqa}. Spatial reference tasks focus on grounding natural-language descriptions in embodied search~\citep{qi2019reverie,zhou2025roborefer}, and manipulation
~\citep{jiang2023vima,mees2022calvin,srivastava2022behavior,wu2023smartplay}. While existing benchmarks incorporate active perception, they largely rely on task-driven foraging. This paradigm conflates the efficiency of environmental exploration with downstream task performance, often fostering brittle spatial representations that lack generalizability~\citep{bonawitz2011double}. Beyond the above task-driven active exploration, EXCALIBUR\cite{zhu2023excalibur} also considers task-agnostic exploration, but its RL training can induce goal leakage and encodes maps implicitly in policy weights. 
In contrast, we study zero-shot foundation-model agents with no environment-specific training for task-agnostic exploration, emphasizing exploration efficiency via minimal-cost uncertainty reduction (rather than coverage), and evaluating not only task success but also the belief construction process via explicit belief probing.

\section{Conclusions}

We introduce \tos, which asks whether foundation models can function as spatial agents under partial observability: not merely answering questions from fixed views, but actively acquiring information through self-directed exploration to \emph{construct}, \emph{revise}, and \emph{exploit} an internal spatial belief. Building on this framing, we contribute a new evaluation paradigm centered on task-agnostic active exploration, downstream spatial tasks for belief exploitation assessment, and explicit probing of internal beliefs via cognitive-map externalization. We implement \tos in a multimodal environment that instantiates parallel text- and vision-based worlds, enabling controlled diagnosis of failures across symbolic versus perceptual observation streams.
A key strength of this design is that it makes spatial belief \emph{measurable} rather than implicit. By requiring models to externalize evolving cognitive maps and uncertainty over unobserved regions, \tos evaluates more than end task accuracy: it reveals the correctness, internal consistency, and temporal dynamics of belief formation, and quantifies how localized mistakes propagate into global map corruption over time. 
Empirically, \textbf{active exploration is a major bottleneck}: end-task performance drops and exploration is less efficient than passive viewing, with the gap widening as room complexity increases. Belief probes make these error sources explicit: in vision, \textbf{perception error} often appears early, and models also exhibit belief \textbf{instability}, where correct information is later overwritten or forgotten, cascading into inconsistencies and lower map fidelity. Finally, when environments change and previously held beliefs must be revised, models exhibit strong \textbf{belief inertia}. They fail to overwrite obsolete priors, and this inertia is especially pronounced for vision-based models, particularly for orientation and facing updates.
Taken together, \tos reframes spatial evaluation from “can the model answer?” to “can the model \emph{build and maintain} a coherent, revisable spatial world model through efficient information gathering?” We hope this benchmark and its belief-centric measurements provide a foundation for developing models with (i) uncertainty-aware and efficient exploration policies, (ii) robust state/belief maintenance under long horizons, and (iii) reliable mechanisms for revising beliefs when the world changes.

\bibliography{main}
\bibliographystyle{main}

\newpage
\appendix
\section*{Appendix}

\section{Technical Details}
\label{sec:appendix}

\subsection{Benchmark Construction}
We expose the ToS world as a \emph{Gym-like} interface~\citep{brockman2016openai}: agents interact in discrete steps under partial observability at a resolution of $384\times384$ to \textbf{construct} and \textbf{revise} an internal spatial belief, which we later \textbf{exploit} in evaluation tasks. Scenes are procedurally generated multi-room layouts on an $N{\times}M$ grid with $n$ named indoor objects (each with integer $(x,y)$ and heading in \{N,E,S,W\}) and a randomized agent spawn pose.
We restrict multi-room layouts to a tree topology: the room–adjacency graph is connected and acyclic (no loops).

\begin{figure}[ht]
    \centering
    \begin{subfigure}[t]{0.4\linewidth}
        \centering
        \includegraphics[width=\linewidth]{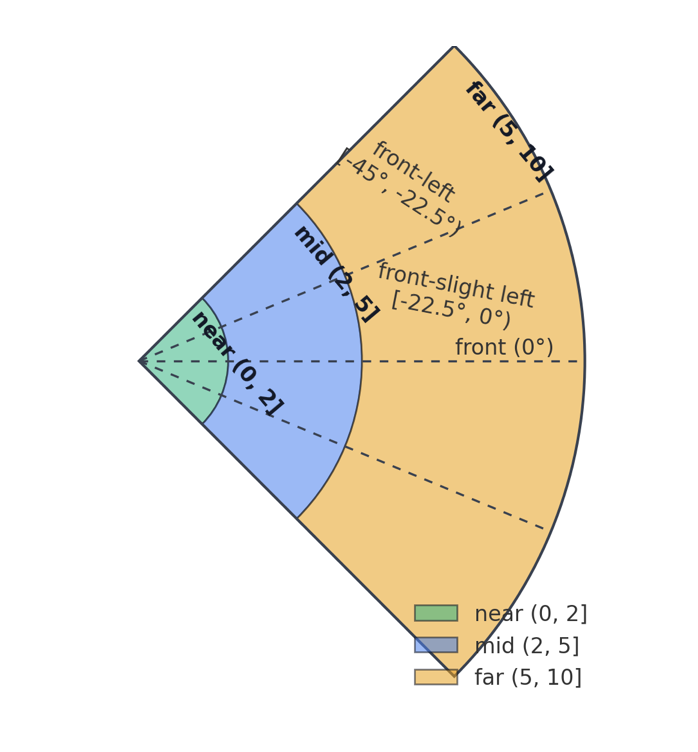}
        \caption{\textbf{Field of view (FOV) specification for the agent in our tasks.} The FOV spans 90° in front of the agent and is divided into angular bins (e.g., front, front-slight left, front-left) and distance ranges (near [0,2], mid [2,5], far [5,10]). This egocentric perception defines how spatial relations are observed and reported.}
        \label{fig:field_of_view}
    \end{subfigure}
    \hfill
    \begin{subfigure}[t]{0.48\linewidth}
        \centering
        \includegraphics[width=\linewidth]{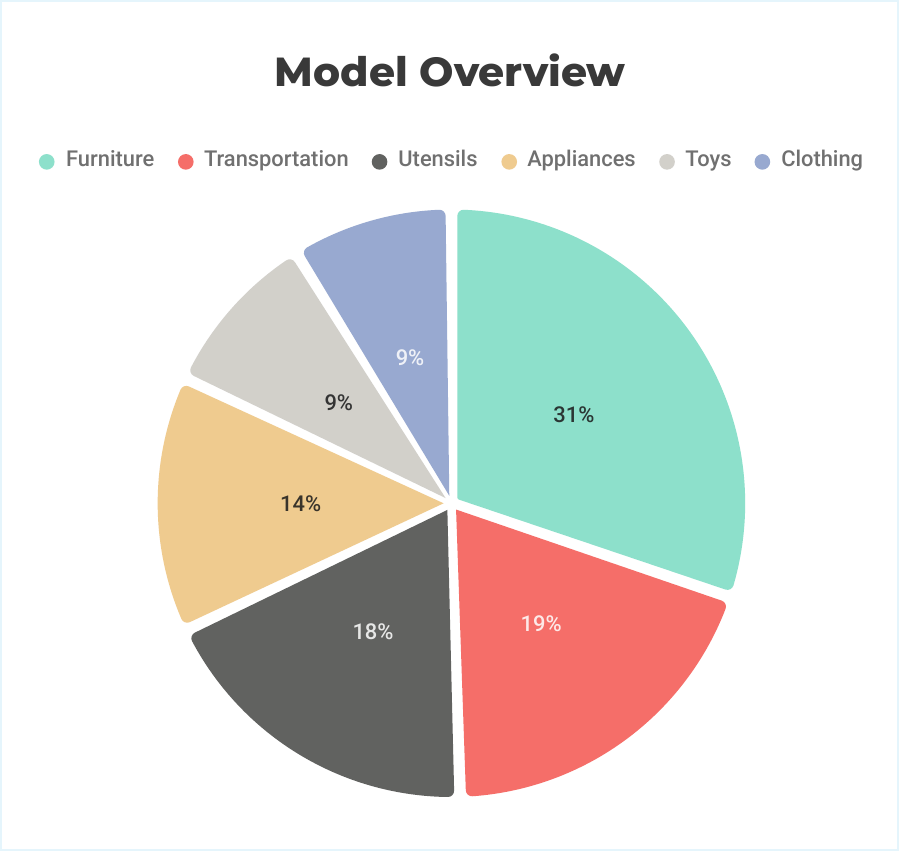}
        \caption{\textbf{Distribution of all 3D models used in our vision tasks.}}
        \label{fig:model_overview}
    \end{subfigure}
    \captionsetup{justification=centering}
    \caption{\textbf{Demonstration figures for FOV and 3D model distribution}}
\end{figure}

\paragraph{Text-based World} At each step, \textsc{Observe} returns a symbolic snapshot of objects in the current room within a $90^\circ$ forward FOV. For every visible object we provide discretized egocentric direction (e.g., \textit{front-left}) and distance bins (e.g., \textit{near}/\textit{mid}/\textit{far}), plus object identity and facing when determinable.
Egocentric observations are rendered with a $90$-degree field of view (FOV), discretized into angular and distance bins as specified in Figure~\ref{fig:field_of_view}. 
Visibility is room-bounded; doorways act as transparent portals only when the agent stands in them, enabling dual-room visibility. Optional noise modules perturb bins for ablations.

\paragraph{Vision-based World}
\label{sec:app:vision-world}

We procedurally generate scenes in a 3D simulator with two controllable parameters: the level (number of rooms) and the object count per room. Objects are drawn from a library of 293 distinct 3D models, grouped into 6 categories and 37 subtypes, primarily everyday household items (see Figure~\ref{fig:model_overview}). To ensure diversity, each object type appears at most once in a given scene.

\begin{figure}[ht!]
    \centering
\includegraphics[width=0.8\linewidth]{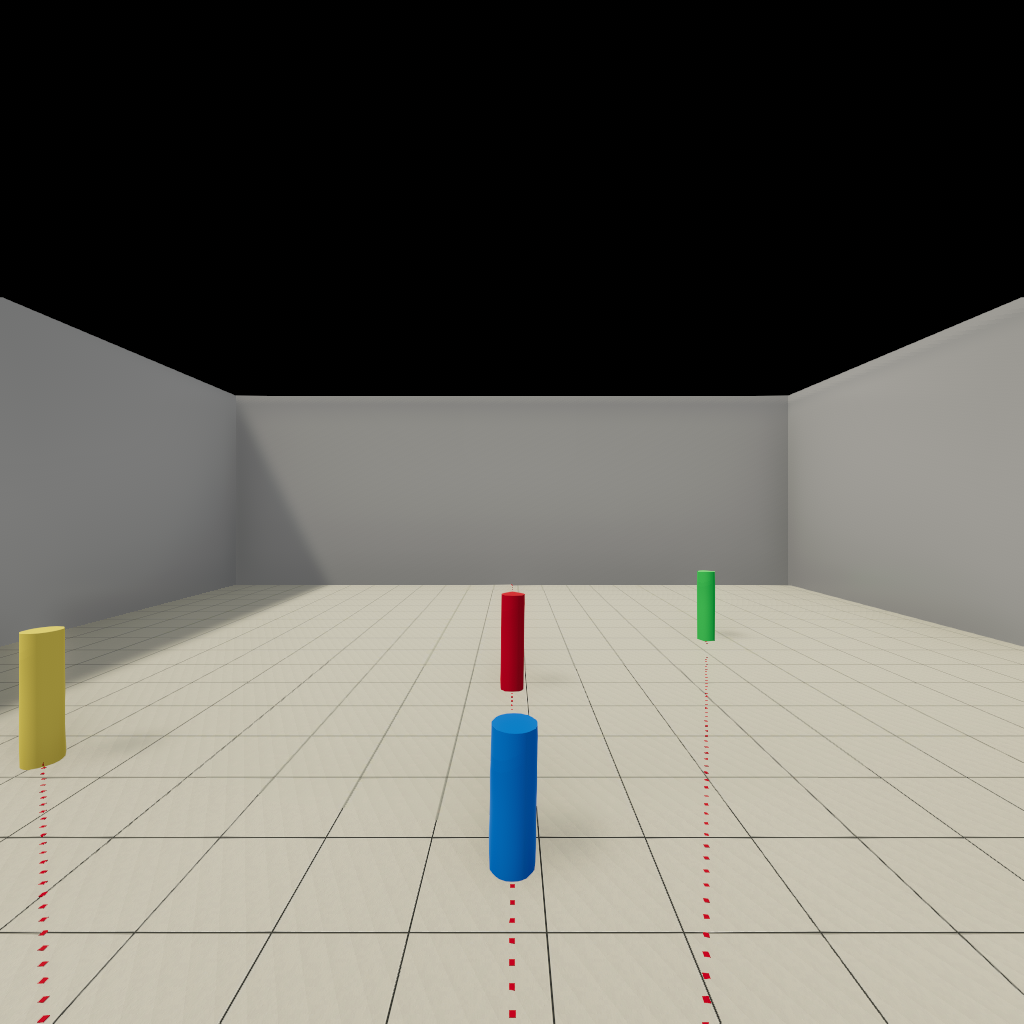}
    \caption{\textbf{Example of distance cues in the vision prompt.} The colored cylinders illustrate objects placed at different distances from the agent: yellow at 2 m, blue at 1 m, red at 2 m, and green at 3 m, providing calibration for mapping visual observations to discretized distance bins.}

    \label{fig:vision_distance_instruction}
\end{figure}

\begin{figure}[ht!]
    \centering
\includegraphics[width=0.8\linewidth]{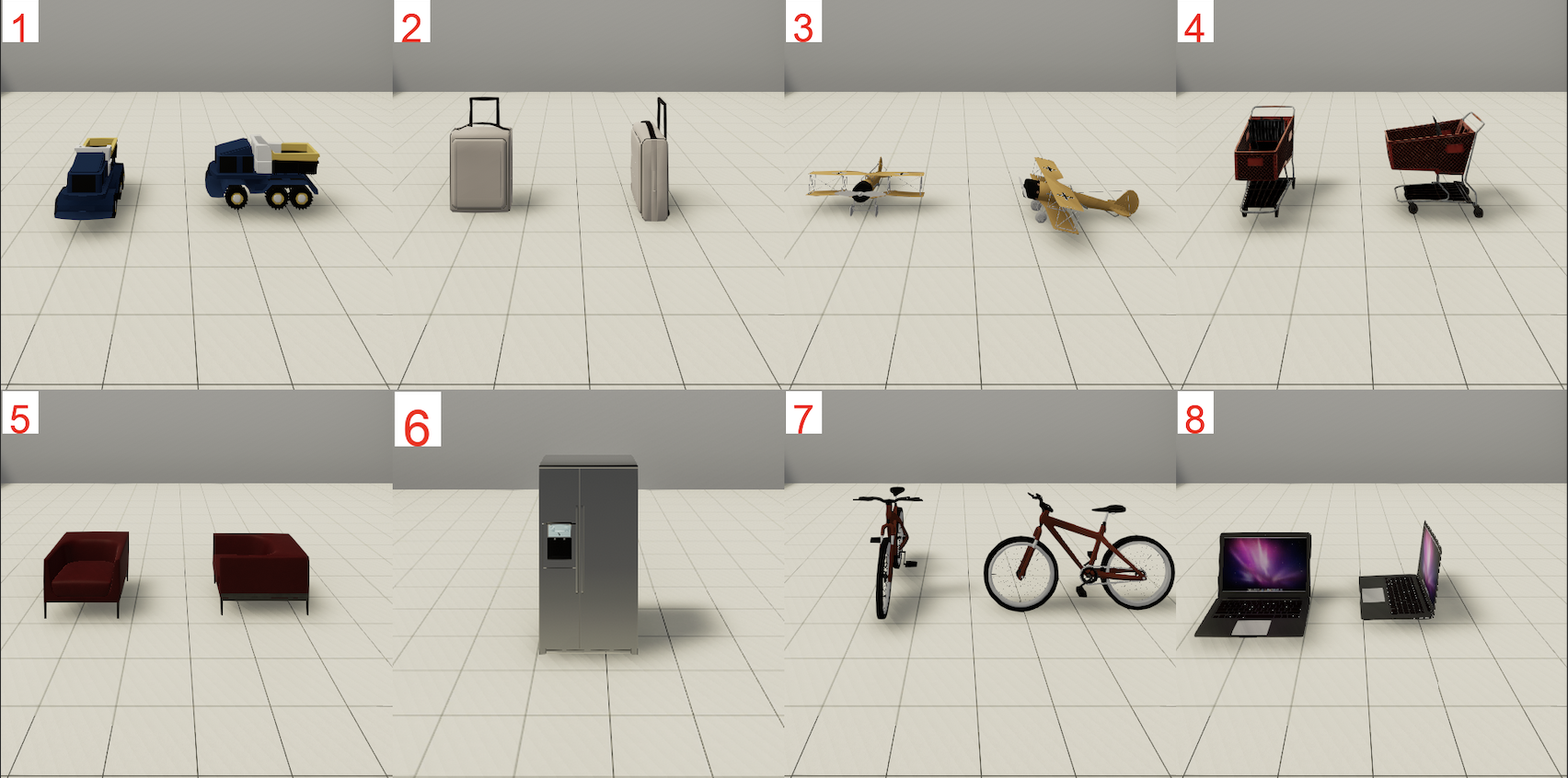}
    \caption{\textbf{Object appearance and orientation cues in the vision prompt.} Objects with facing direction are shown from both the front and side views, while objects without inherent orientation are displayed only from the front view. This provides the agent with consistent visual references for recognizing shape and facing.}

    \label{fig:vision_orientation_instruction}
\end{figure}

For task setup, we additionally generate instructional (Figure~\ref{fig:vision_distance_instruction}) and orientation (Figure~\ref{fig:vision_orientation_instruction}) images that serve as references for the agent in vision-world. We include both images in the vision prompt. Object placement follows validity constraints (e.g., collision avoidance, minimum spacing), and random seeds control reproducibility across environments.

\paragraph{Information Gain Calculation}
We use the AC-3 arc-consistency algorithm to maintain, for each object, a domain of feasible grid cells. Initially, every object’s domain spans the entire $20 \times 20$ map. Each new observation is compiled into unary and binary constraints (e.g., egocentric direction/distance bins, room visibility/occlusion, and \textsc{AllDifferent} to prevent collisions). When a constraint is added, AC-3 iteratively prunes any cell in one object’s domain that is unsupported by the domains of related objects, propagating revisions along incident arcs until a fixed point is reached (all arcs are consistent). While AC-3 alone does not guarantee global consistency, in our setting all constraints are derived from a valid trajectory; therefore the ground-truth assignment remains supported and is never pruned, ensuring that domains stay non-empty throughout propagation.

\paragraph{Proxy agents} 
\label{par:proxy}
We implement two scripted proxies to provide strong, reproducible baselines.

\textbf{\textsc{Scout}.} From its spawn pose, the agent performs a 360° sweep (four cardinal \textsc{Rotate}+\textsc{Observe} actions) to capture all views at the initial location. It then follows a fixed room-visitation order: upon discovering a doorway, it enters the adjacent room, executes the same sequential sweep, and repeats this “visit–sweep–advance” routine until every room has been observed at least once.

\textbf{\textsc{Strategist}.} The first stage mirrors \textsc{Scout}: a panoramic sweep to register all currently visible objects. Thereafter, within the \emph{current room} the agent maintains, for each object, a set of feasible positions (“domain”) induced by accumulated observations. At each turn it:
(i) selects the object with the largest remaining domain (highest positional uncertainty);
(ii) moves to a viewpoint that best constrains this object (e.g., near it or along a sightline that intersects the most candidate cells);
(iii) at that viewpoint, orients to test pairwise relations: it computes unresolved pairwise directions between the target object and all others in the room, identifies the direction bin with the highest outstanding count, and \textsc{Observe}s in that orientation first.
The procedure iterates until all objects in the room are resolved (domains shrink to singletons), then proceeds to the next unvisited room and repeats.

\paragraph{Prompts}
\label{sec:app:prompts}

\begin{figure}[ht!]
    \centering
\includegraphics[width=\linewidth]{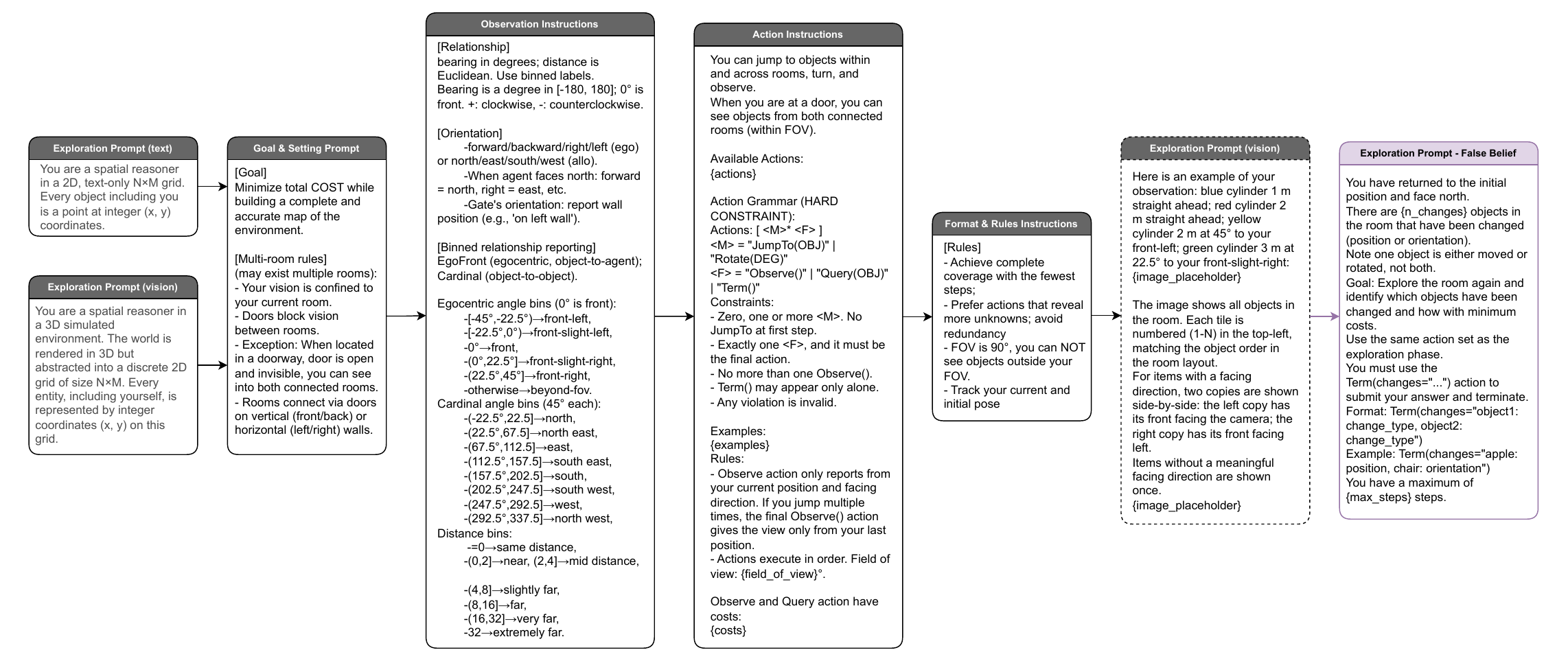}
    \caption{\textbf{Exploration prompts}}

    \label{fig:exploration_prompts}
\end{figure}

\begin{figure}[ht!]
    \centering
\includegraphics[width=\linewidth]{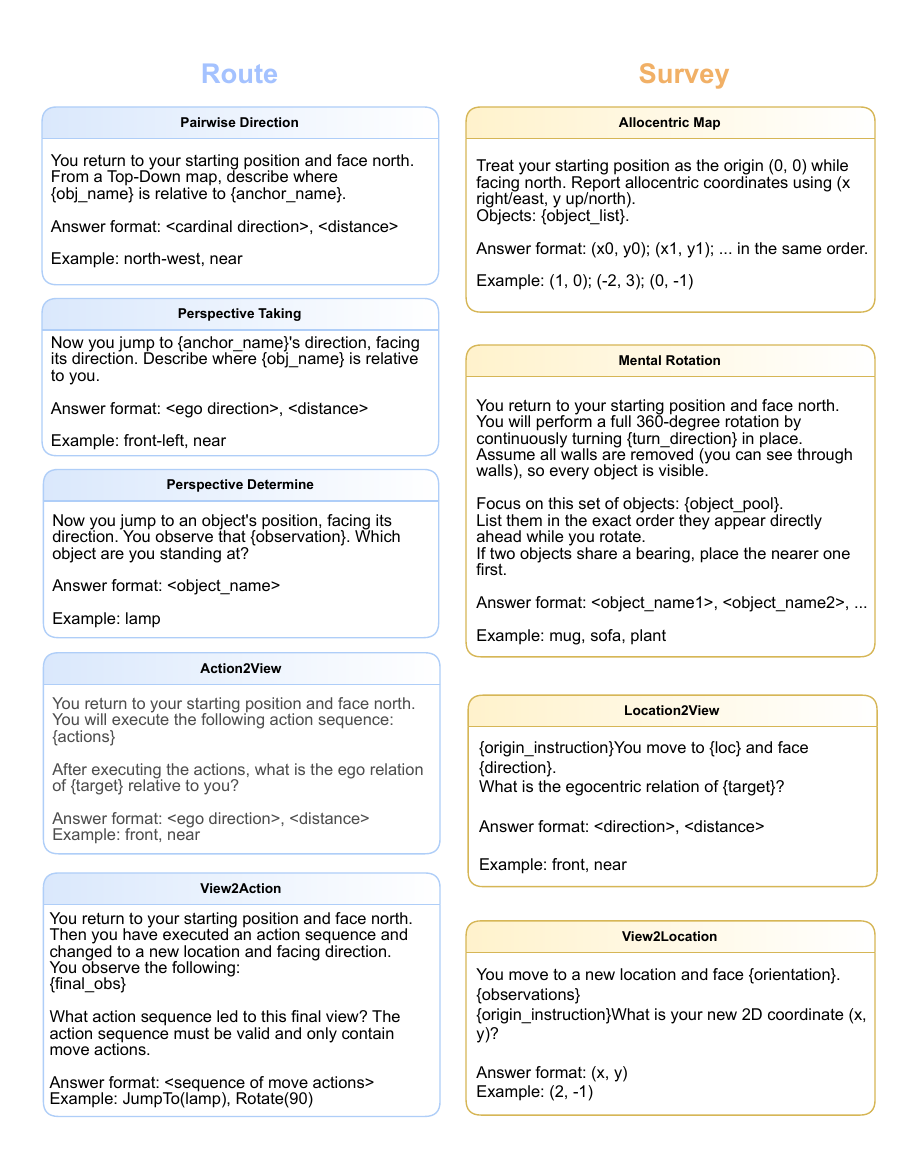}
    \caption{\textbf{Evaluation prompt design. We show the prompt for each evaluation task.}}

    \label{fig:evaluation_prompts}
\end{figure}

\begin{figure}[ht!]
    \centering
\includegraphics[width=\linewidth]{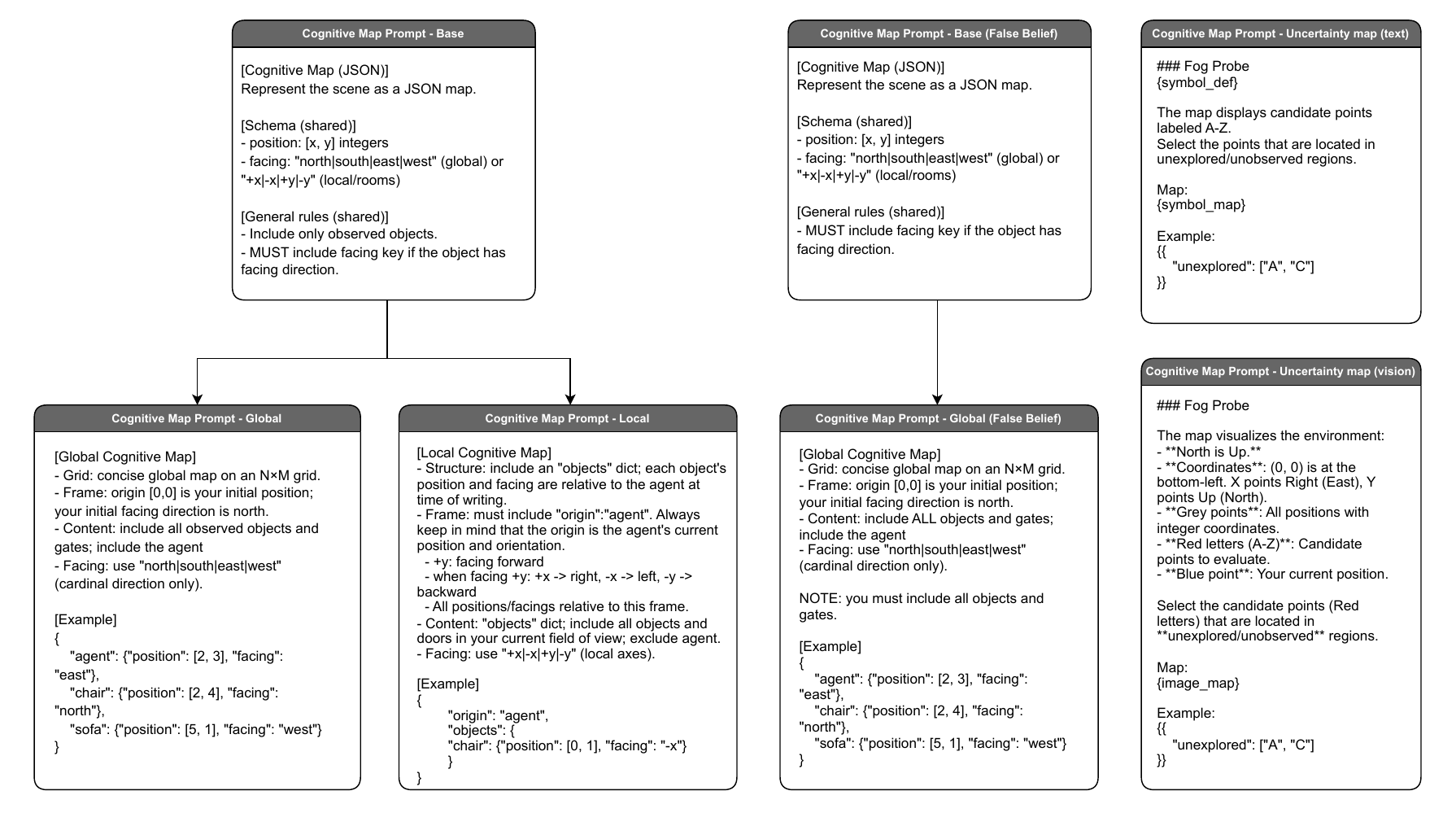}
    \caption{\textbf{Belief probing prompt design. We use these prompts to ask the model to output a cognitive map or select unobserved points.}}

    \label{fig:cognitive_prompts}
\end{figure}

\begin{figure}[ht!]
    \centering
\includegraphics[width=\linewidth]{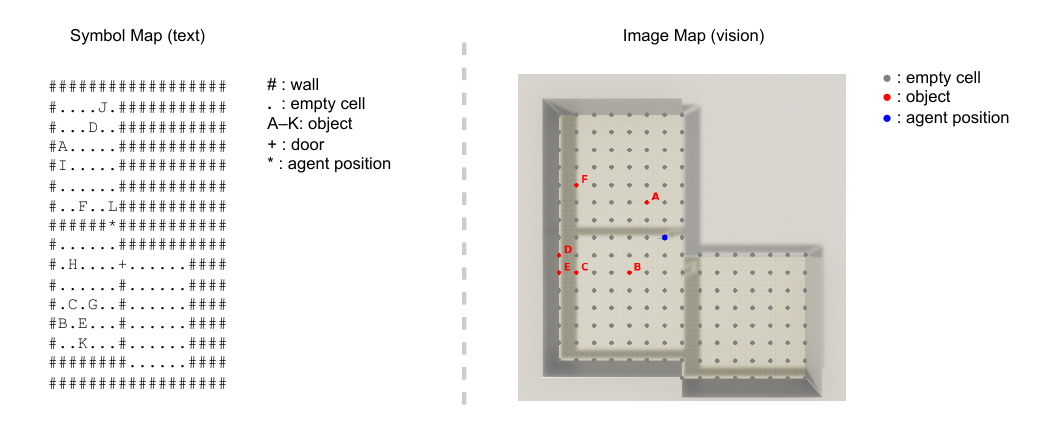}
    \caption{\textbf{The symbol map and the image map}
provide parallel representations of the same environment for text and vision settings in uncertainty probing prompts.}
    \label{fig:symbol_map}
\end{figure}

We show the detailed designs of our prompts for exploration in Figure~\ref{fig:exploration_prompts}, evaluation prompts in Figure~\ref{fig:evaluation_prompts}, cognitive map prompts in Figure~\ref{fig:cognitive_prompts}, and top-down view for uncertainty modeling in Figure~\ref{fig:symbol_map}.

\section{Evaluation Setups}
\label{app:eval}
% step
% action fail handling
% 
To enable a like-for-like comparison between the text and vision settings, we instantiate identical room layouts across modalities. Concretely, we generate $100$ evaluation instances with IDs $0–99$; for each ID, we use the ID itself as the random seed to drive task sampling in both environments. This seed tying guarantees deterministic layouts and bit-for-bit reproducibility across modalities.

\paragraph{Additional Results}
\label{par:additional_results}
% other room settings
We show detailed results for \textbf{different room settings} including two-room and four-room layouts. 
In both the two-room and four-room settings, we use the same room size and the same number of objects per room as in the three-room setting. For the four-room setting, we connect the main room with all the others. 
We evaluate \textsc{GPT-5.2} and \textsc{Gemini-3 Pro}, the two best-performing models. \textbf{Additionally, we tested higher resolution, but found no performance gain.}
Table~\ref{tab:passive-2} and \ref{tab:active-2} report passive and active performance of the two-room setting. Table~\ref{tab:passive-3} and \ref{tab:active-3} report passive and active performance of the three-room setting. As the number of rooms increases, exploration cost rises accordingly. The results also underscore the importance of efficient exploration: in the four-room setting, which demands more strategic exploration, the gap between active and passive performance becomes substantially larger.

\begin{center}
\small
\setlength{\tabcolsep}{6pt}
\setlength\extrarowheight{1pt}
\resizebox{0.85\textwidth}{!}{%
\begin{tabular}{l *{8}{C{\rescolwexp}} C{1.4\rescolwexp} @{} >{\columncolor{white}}m{0.35cm} !{\vrule width 0.5pt} @{} C{1.00cm}}

% ---- TOP: task names (now above Exploration/Route/Survey) ----
\multicolumn{1}{c}{} &
\roth{direction} & \roth{persp.take} & \roth{perc.dec} & \roth{act2view} & \roth{view2act} &
\roth{alloc.map} & \roth{ment.rot} & \roth{loc2view} & \roth{view2loc} &
\multicolumn{1}{c}{} & \multicolumn{1}{c}{}\\

% ---- Level 1: Static / Dynamic (colored labels only) ----
\multicolumn{1}{c}{} &
\multicolumn{1}{c}{\cellcolor{statbg}\textit{Static (S)}} &
\multicolumn{4}{c}{\cellcolor{dynbg}\textit{Dynamic (D)}} &
\multicolumn{1}{c}{\cellcolor{statbg}\textit{Static (S)}} &
\multicolumn{3}{c}{\cellcolor{dynbg}\textit{Dynamic (D)}} &
\multicolumn{1}{c}{} & \multicolumn{1}{c}{}\\

% ---- Level 2: Exploration / Route / Survey (colored labels only) ----
\textbf{Methods} &
\multicolumn{5}{c}{\cellcolor{routebg}\textbf{Route}} &
\multicolumn{4}{c}{\cellcolor{surveybg}\textbf{Survey}} &
\multicolumn{1}{c}{} & \textbf{Avg.}\\
\hline

\multicolumn{11}{c}{Vision-based World}\\
\hline
\rowcolor{secgray}\multicolumn{11}{l}{\textit{Proprietary Models}} \\
\textsc{GPT-5.2} & 39.2 & 37.3 & 63.3 & 53.8 & 58.3 & 68.2 & 92.7 & 52.3 & 68.6 && 59.3 \\
\textsc{Gemini-3 Pro} & 57.8 & 33.9 & 53.8 & 48.5 & 58.7 & 64.6 & 83.3 & 54.7 & 69.8 && 58.3 \\
\hline

\multicolumn{11}{c}{Text-based World}\\
\hline
\rowcolor{secgray}\multicolumn{11}{l}{\textit{Proprietary Models}} \\
\textsc{GPT-5.2} & 85.3 & 92.0 & 99.0 & 90.0 & 83.0 & 97.2 & 99.7 & 89.5 & 95.2 && 92.3 \\
\textsc{Gemini-3 Pro} & 88.2 & 86.7 & 91.7 & 87.3 & 79.3 & 90.1 & 92.7 & 81.5 & 82.9 && 86.7 \\
\hline

\end{tabular}
}
\captionof{table}{Exploitation Performance ($\%$) via Passive Observations under \textbf{two rooms} settings.}

\label{tab:passive-2}
\end{center}

\begin{center}
\small
\setlength{\tabcolsep}{6pt}
\setlength\extrarowheight{1pt}
\resizebox{0.85\textwidth}{!}{%
\begin{tabular}{l C{1.4\rescolwexp} *{8}{C{0.92\rescolwexp}} C{1.5\rescolwexp} @{} >{\columncolor{white}}m{0.35cm} !{\vrule width 0.5pt} @{} C{1.00cm}}

% ---- TOP: task names (now above Exploration/Route/Survey) ----
\multicolumn{1}{c}{} &  &
\roth{direction} & \roth{persp.take} & \roth{perc.dec.} & \roth{act2view} & \roth{view2act} &
\roth{alloc.map} & \roth{ment.rot} & \roth{loc2view} & \roth{view2loc} &
\multicolumn{1}{c}{} & \multicolumn{1}{c}{}\\

% ---- Level 1: Static / Dynamic (colored labels only) ----
\multicolumn{1}{c}{} &
\multicolumn{1}{c}{} % Avg.cost: no S/D labels
&
\multicolumn{1}{c}{\cellcolor{statbg}\textit{Static (S)}} &
\multicolumn{4}{c}{\cellcolor{dynbg}\textit{Dynamic (D)}} &
\multicolumn{1}{c}{\cellcolor{statbg}\textit{Static (S)}} &
\multicolumn{3}{c}{\cellcolor{dynbg}\textit{Dynamic (D)}} &
\multicolumn{1}{c}{} & \multicolumn{1}{c}{}\\

% ---- Level 2: Exploration / Route / Survey (colored labels only) ----
\textbf{Methods} &
\multicolumn{1}{c}{\cellcolor{surveybg}\textbf{Avg.cost}}
&
\multicolumn{5}{c}{\cellcolor{routebg}\textbf{Route}} &
\multicolumn{4}{c}{\cellcolor{surveybg}\textbf{Survey}} &
\multicolumn{1}{c}{} & \textbf{Avg.}\\
\hline

\multicolumn{12}{c}{Vision-based World}\\
\hline
\rowcolor{secgray}\multicolumn{12}{l}{\textit{Proprietary Models}} \\
\textsc{GPT-5.2} & 10.8 & 41.3 & 36.2 & 48.2 & 49.0 & 54.7 & 56.9 & 72.0 & 45.2 & 59.7 & & 51.5 \\
\textsc{Gemini-3 Pro} & 6.6 & 51.7 & 36.3 & 63.0 & 47.2 & 56.0 & 63.4 & 85.0 & 50.3 & 67.5 & & 57.8 \\
\hline

\multicolumn{12}{c}{Text-based World}\\
\hline
\rowcolor{secgray}\multicolumn{12}{l}{\textit{Proprietary Models}} \\
\textsc{GPT-5.2} & 6.2 & 68.7 & 67.3 & 90.0 & 76.8 & 64.0 & 83.4 & 92.7 & 73.7 & 83.7 & & 77.8 \\
\textsc{Gemini-3 Pro} & 6.2 & 76.0 & 68.3 & 89.0 & 77.2 & 72.7 & 83.1 & 96.0 & 77.5 & 86.2 & & 80.6 \\
\hline

\end{tabular}
}
\captionof{table}{Exploitation Performance ($\%$) via Active Exploration under \textbf{two rooms} settings.}
\label{tab:active-2}
\end{center}

\begin{center}
\small
\setlength{\tabcolsep}{6pt}
\setlength\extrarowheight{1pt}
\resizebox{0.85\textwidth}{!}{%
\begin{tabular}{l *{8}{C{\rescolwexp}} C{1.4\rescolwexp} @{} >{\columncolor{white}}m{0.35cm} !{\vrule width 0.5pt} @{} C{1.00cm}}

% ---- TOP: task names (now above Exploration/Route/Survey) ----
\multicolumn{1}{c}{} &
\roth{direction} & \roth{persp.take} & \roth{perc.dec} & \roth{act2view} & \roth{view2act} &
\roth{alloc.map} & \roth{ment.rot} & \roth{loc2view} & \roth{view2loc} &
\multicolumn{1}{c}{} & \multicolumn{1}{c}{}\\

% ---- Level 1: Static / Dynamic (colored labels only) ----
\multicolumn{1}{c}{} &
\multicolumn{1}{c}{\cellcolor{statbg}\textit{Static (S)}} &
\multicolumn{4}{c}{\cellcolor{dynbg}\textit{Dynamic (D)}} &
\multicolumn{1}{c}{\cellcolor{statbg}\textit{Static (S)}} &
\multicolumn{3}{c}{\cellcolor{dynbg}\textit{Dynamic (D)}} &
\multicolumn{1}{c}{} & \multicolumn{1}{c}{}\\

% ---- Level 2: Exploration / Route / Survey (colored labels only) ----
\textbf{Methods} &
\multicolumn{5}{c}{\cellcolor{routebg}\textbf{Route}} &
\multicolumn{4}{c}{\cellcolor{surveybg}\textbf{Survey}} &
\multicolumn{1}{c}{} & \textbf{Avg.}\\
\hline

\multicolumn{11}{c}{Vision-based World}\\
\hline
\rowcolor{secgray}\multicolumn{11}{l}{\textit{Proprietary Models}} \\
\textsc{GPT-5.2} & 47.0 & 37.7 & 59.7 & 38.3 & 40.3 & 60.1 & 73.7 & 50.5 & 65.9 && 52.6 \\
\textsc{Gemini-3 Pro} & 63.5 & 35.5 & 58.7 & 42.8 & 43.0 & 64.4 & 81.7 & 48.8 & 67.4 && 56.2 \\
\hline
\multicolumn{11}{c}{Text-based World}\\
\hline
\rowcolor{secgray}\multicolumn{11}{l}{\textit{Proprietary Models}} \\
\textsc{GPT-5.2} & 83.8 & 88.2 & 94.3 & 86.8 & 62.7 & 94.8 & 93.7 & 82.0 & 92.5 && 86.5 \\
\textsc{Gemini-3 Pro} & 81.2 & 91.3 & 96.7 & 82.2 & 68.3 & 76.8 & 81.3 & 74.2 & 79.0 && 81.2 \\
\hline

\end{tabular}
}
\captionof{table}{Exploitation Performance ($\%$) via Passive Observations under \textbf{four rooms} settings.}

\label{tab:passive-3}
\end{center}

\begin{center}
\small
\setlength{\tabcolsep}{6pt}
\setlength\extrarowheight{1pt}
\resizebox{0.85\textwidth}{!}{%
\begin{tabular}{l C{1.4\rescolwexp} *{8}{C{0.92\rescolwexp}} C{1.5\rescolwexp} @{} >{\columncolor{white}}m{0.35cm} !{\vrule width 0.5pt} @{} C{1.00cm}}

% ---- TOP: task names (now above Exploration/Route/Survey) ----
\multicolumn{1}{c}{} &  &
\roth{direction} & \roth{persp.take} & \roth{perc.dec.} & \roth{act2view} & \roth{view2act} &
\roth{alloc.map} & \roth{ment.rot} & \roth{loc2view} & \roth{view2loc} &
\multicolumn{1}{c}{} & \multicolumn{1}{c}{}\\

% ---- Level 1: Static / Dynamic (colored labels only) ----
\multicolumn{1}{c}{} &
\multicolumn{1}{c}{} % Avg.cost: no S/D labels
&
\multicolumn{1}{c}{\cellcolor{statbg}\textit{Static (S)}} &
\multicolumn{4}{c}{\cellcolor{dynbg}\textit{Dynamic (D)}} &
\multicolumn{1}{c}{\cellcolor{statbg}\textit{Static (S)}} &
\multicolumn{3}{c}{\cellcolor{dynbg}\textit{Dynamic (D)}} &
\multicolumn{1}{c}{} & \multicolumn{1}{c}{}\\

% ---- Level 2: Exploration / Route / Survey (colored labels only) ----
\textbf{Methods} &
\multicolumn{1}{c}{\cellcolor{surveybg}\textbf{Avg.cost}}
&
\multicolumn{5}{c}{\cellcolor{routebg}\textbf{Route}} &
\multicolumn{4}{c}{\cellcolor{surveybg}\textbf{Survey}} &
\multicolumn{1}{c}{} & \textbf{Avg.}\\
\hline

\multicolumn{12}{c}{Vision-based World}\\
\hline
\rowcolor{secgray}\multicolumn{12}{l}{\textit{Proprietary Models}} \\
\textsc{GPT-5.2} & 23.2 & 41.2 & 33.2 & 49.0 & 30.8 & 30.7 & 32.5 & 49.7 & 40.5 & 55.4 & & 40.3 \\
\textsc{Gemini-3 Pro} & 19.7 & 59.8 & 34.2 & 60.3 & 34.7 & 46.0 & 56.8 & 62.7 & 44.0 & 64.8 & & 51.5 \\
\hline

\multicolumn{12}{c}{Text-based World}\\
\hline
\rowcolor{secgray}\multicolumn{12}{l}{\textit{Proprietary Models}} \\
\textsc{GPT-5.2} & 16.4 & 65.3 & 69.0 & 74.3 & 62.8 & 44.3 & 66.6 & 76.3 & 57.5 & 77.8 & & 66.0 \\
\textsc{Gemini-3 Pro} & 19.7 & 76.3 & 77.2 & 91.7 & 73.3 & 64.3 & 77.0 & 83.7 & 74.0 & 81.9 & & 77.7 \\
\hline

\end{tabular}
}
\captionof{table}{Exploitation Performance ($\%$) via Active Exploration under \textbf{four rooms} settings.}
\label{tab:active-3}
\end{center}

\section{Additional Visualization Examples}
\label{sec:app:examples}

\begin{figure}[ht!]
    \centering
\includegraphics[width=0.9\linewidth]{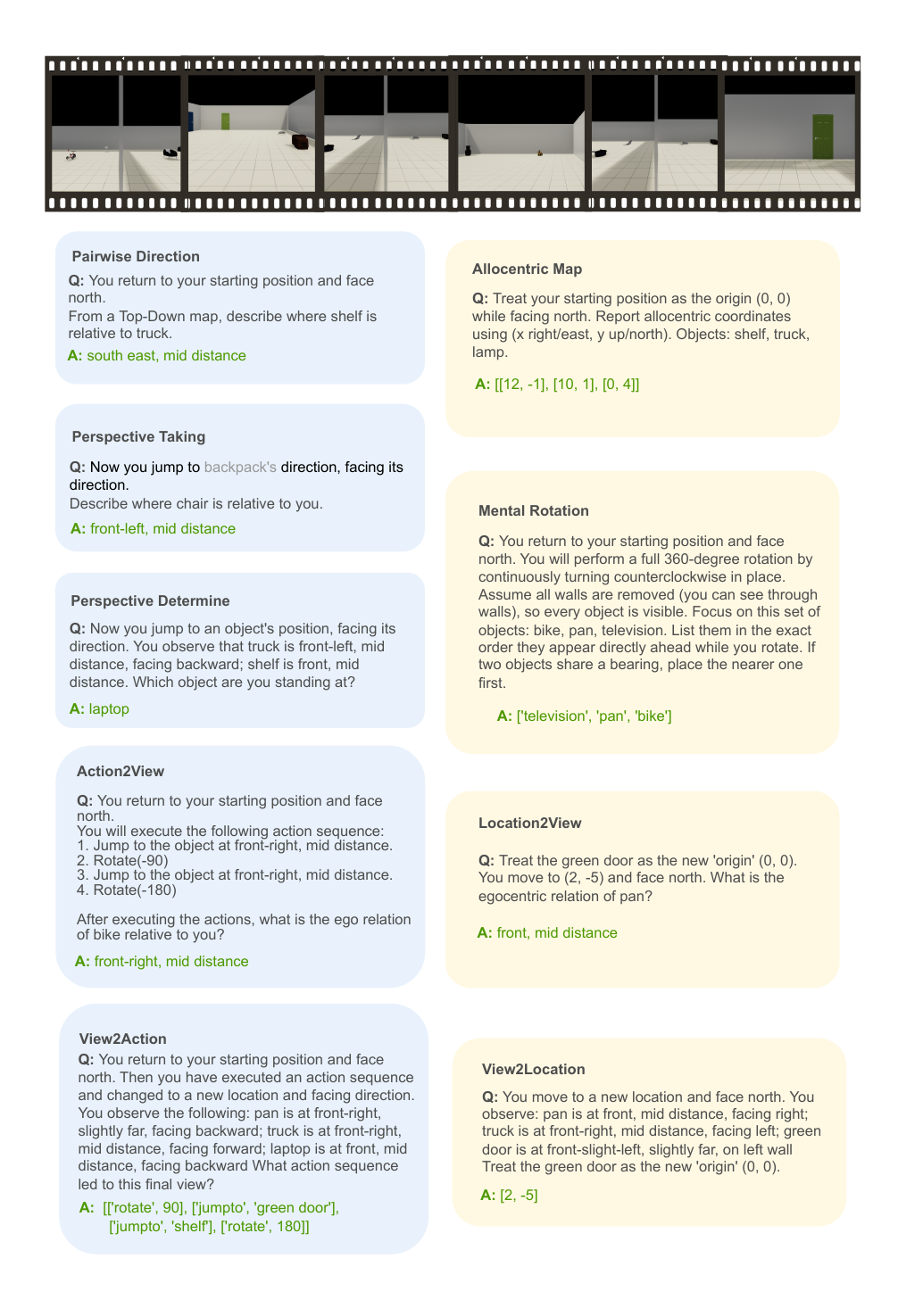}
    \caption{\textbf{Examples of task formats and answer styles used.} Each block illustrates a spatial reasoning task type in our suite (Route-level and Survey-level), including the corresponding input context and an example open-ended answer that must follow a strict output format. In the vision setting, textual scene descriptions in the questions are replaced by rendered observation images.}

    \label{fig:spatial_QA_example}
\end{figure}

We include concrete examples of task formats and answer styles with open-ended, format-constrained outputs in Figure~\ref{fig:spatial_QA_example}.

\begin{figure}[!htbp]
    \centering
    \includegraphics[width=\textwidth]{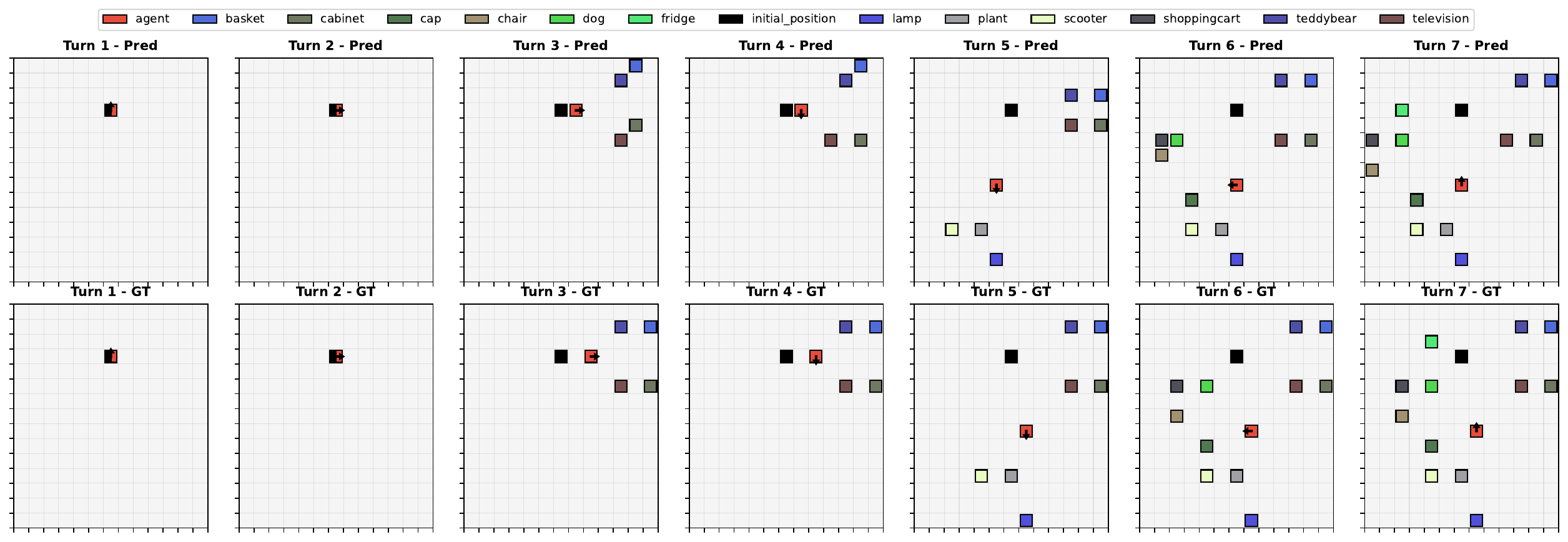}
    \caption{\textsc{GPT-5.2}'s turn-by-turn cognitive map in text world during exploration.
    }
    \label{fig:gpt5_heatmap_cogmap_text}
\end{figure}

\begin{figure}[!htbp]
    \centering
    \includegraphics[width=\textwidth]{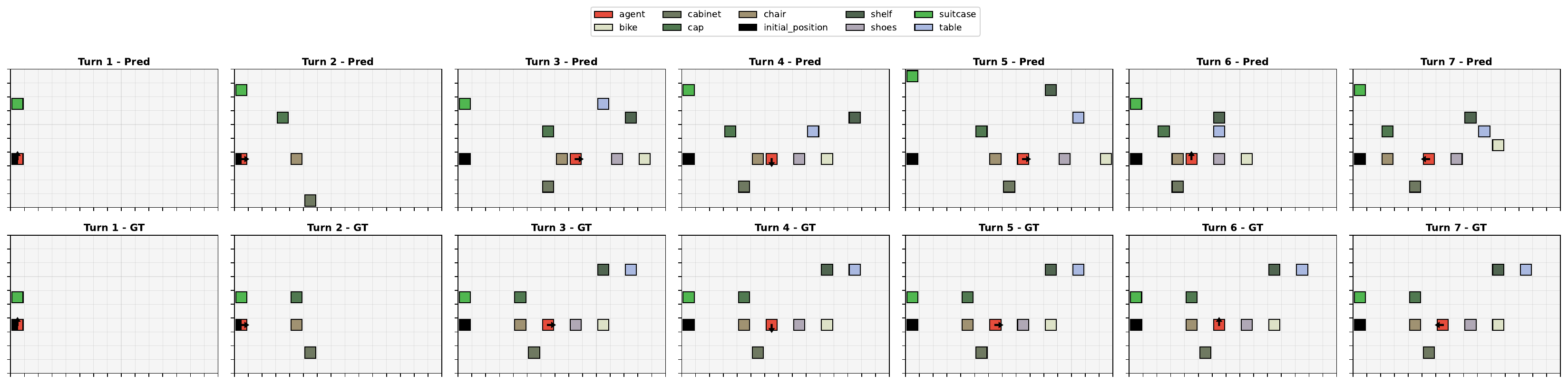}
    \caption{\textsc{GPT-5.2}'s turn-by-turn cognitive map in vision world during exploration.
    }
    \label{fig:gpt5_heatmap_cogmap_vision}
\end{figure}

\paragraph{Cognitive map output by models}
We visualize the turn-by-turn cognitive maps (in Figures~\ref{fig:gpt5_heatmap_cogmap_text} and \ref{fig:gpt5_heatmap_cogmap_vision} of \textsc{GPT-5.2}, comparing them against ground-truth maps. The performance is noticeably stronger in text-based environments than in vision-based ones.

\begin{figure}[!htbp]
    \centering
    \includegraphics[width=\textwidth]{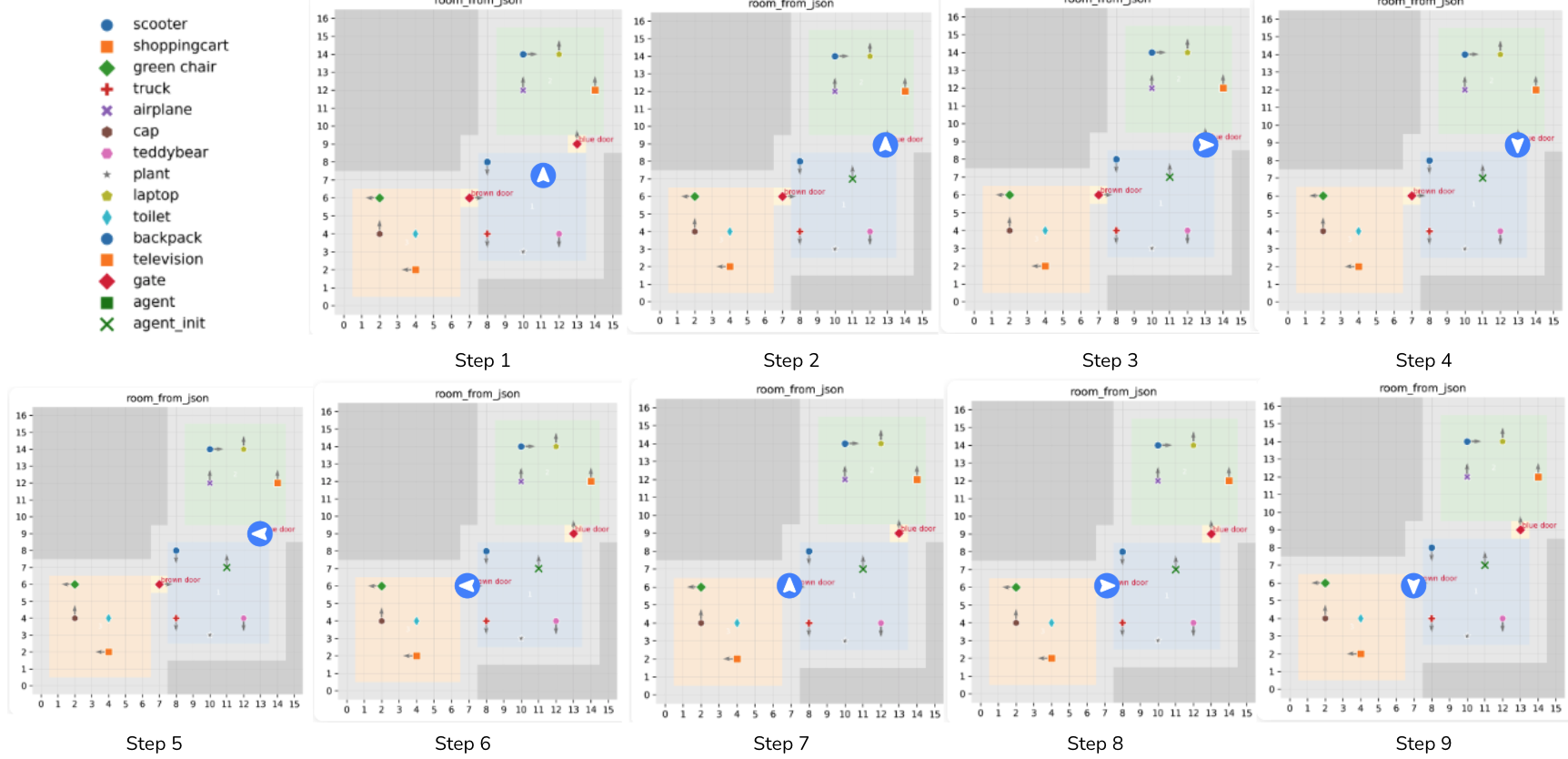}
    \caption{Example trajectory illustrating \textsc{GPT-5.2}’s door-finding strategy and systematic sweeping pattern: Upon detecting a door, the agent navigates toward it and executes a strategic rotation to maximize environmental coverage. The process terminates once all target objects have been successfully identified.
    }
    \label{fig:gpt_systematic_sweeping}
\end{figure}

\begin{figure}[!htbp]
    \centering
    \includegraphics[width=\textwidth]{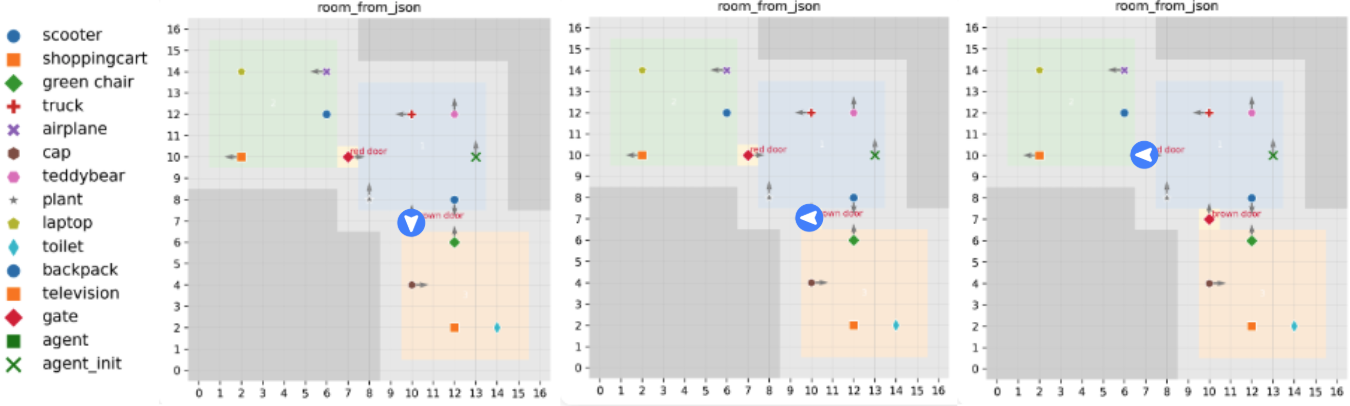}
    \caption{Example trajectory illustrating \textsc{GPT-5.2}’s omission pattern: Observing the door too early may lead the agent to skip the rest of the exploration, causing incomplete environmental discovery. 
    }
    \label{fig:gpt_omission}
\end{figure}

\begin{figure}[!htbp]
    \centering
    \includegraphics[width=\textwidth]{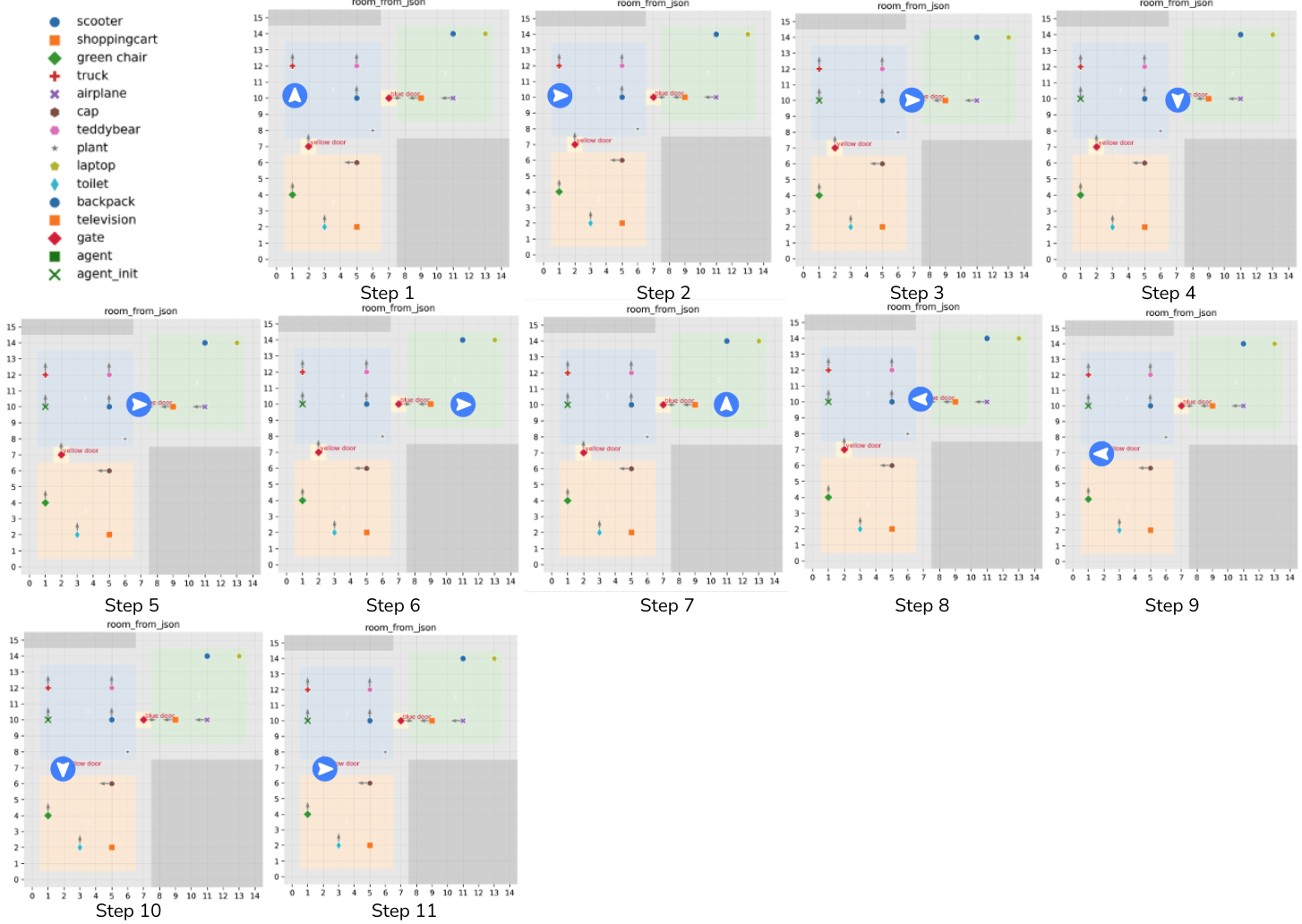}
    \caption{Example trajectory illustrating \textsc{GEMINI-3 Pro}’s door-finding strategy and systematic sweeping pattern in vision world: Upon detecting a door, the agent navigates toward it and executes a strategic rotation to maximize environmental coverage. The process terminates once all target objects have been successfully identified.
    }
    \label{fig:gemini_systematic_sweeping}
\end{figure}

\begin{figure}[!htbp]
    \centering
    \includegraphics[width=\textwidth]{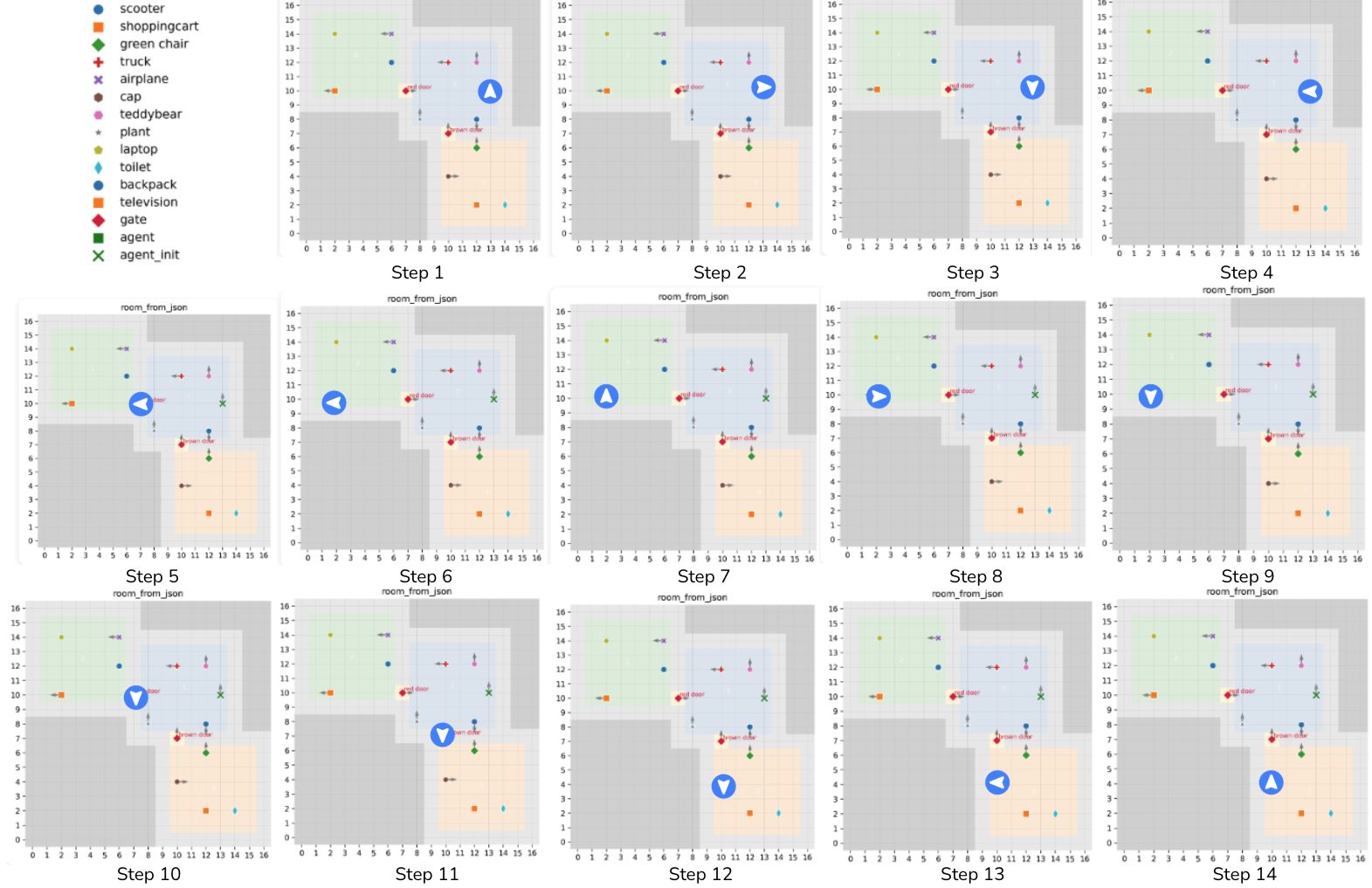}
    \caption{Example trajectory illustrating \textsc{GEMINI-3 Pro}’s object sweeping pattern mostly found in text world: Orbit the starting object using it as the pivot point. Randomly select an observed door to jump to a new object, then resume pivoting around the new target in a continuous loop.
    }
    \label{fig:gemini_object_sweeping}
\end{figure}

\begin{figure}[!htbp]
    \centering
    \includegraphics[width=\textwidth]{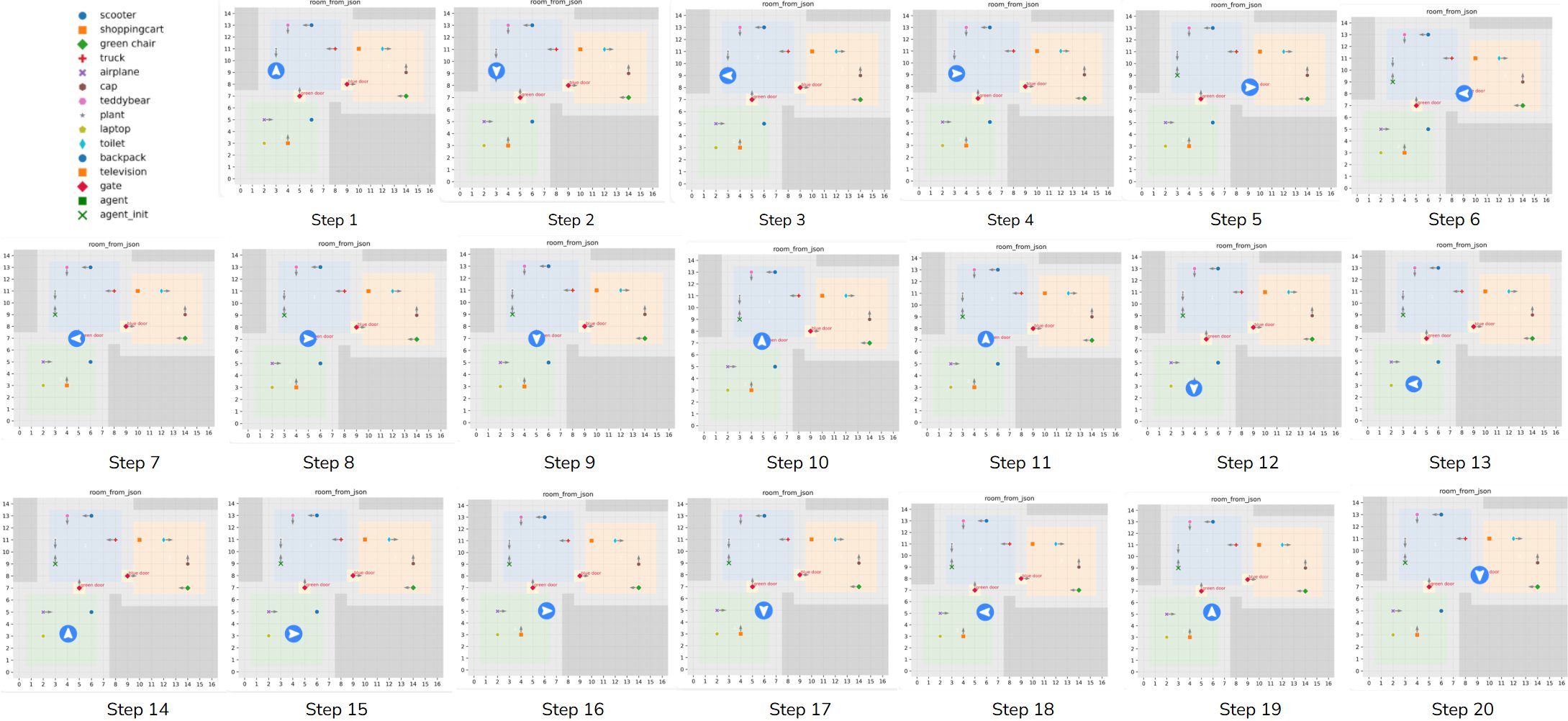}
    \caption{Example trajectory illustrating \textsc{CLAUDE-4.5 Sonnet}’s exploration pattern: There is no clear exploration pattern.
    }
    \label{fig:claude_explore_pattern}
\end{figure}

\paragraph{Exploration pattern examples by models}
\label{par:exp_pattern}
We include representative trajectories from each model to illustrate the active exploration patterns identified in our analysis, shown in Figure~\ref{fig:gpt_systematic_sweeping}, \ref{fig:gpt_omission}, \ref{fig:gemini_systematic_sweeping}, \ref{fig:gemini_object_sweeping}, and~\ref{fig:claude_explore_pattern} . These examples highlight how different models manifest recurring exploration behaviors: for instance, \textsc{GPT-5.2} often adopts a “finding-gate” strategy, rotating until a doorway is detected before moving toward it, while other models more frequently repeat redundant checks. All figures mark the agent’s position and orientation explicitly, with actions annotated beneath each frame and a shared legend provided for each trajectory.

\begin{figure}[!htbp]
    \centering
    \includegraphics[width=\textwidth]{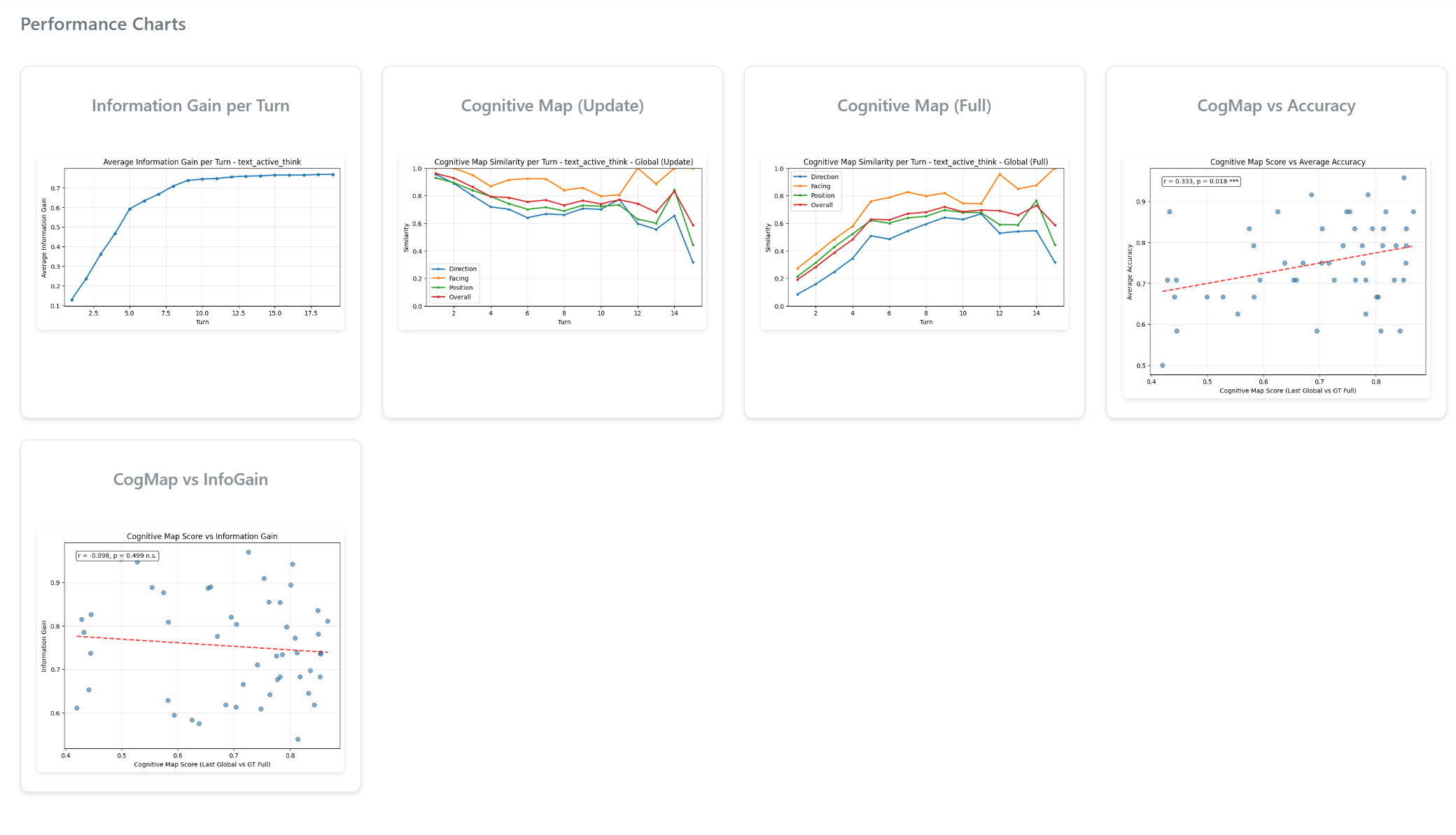}
    \caption{Platform designed by us for analysis (chart)
    }
    \label{fig:ui_chart}
\end{figure}

\begin{figure}[!htbp]
    \centering
    \includegraphics[width=\textwidth]{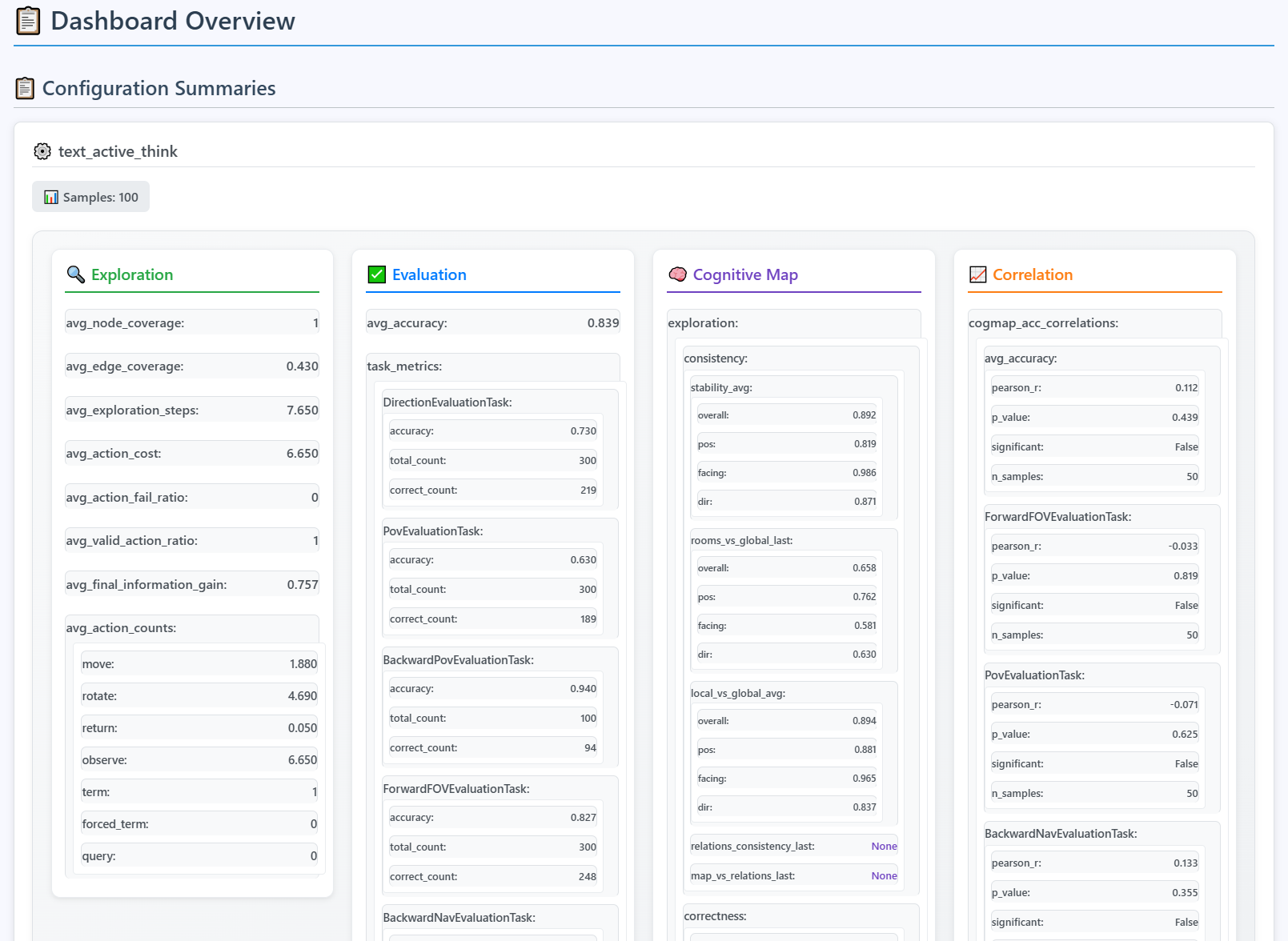}
    \caption{Visualization Platform for analysis: Metrics for active exploration in text world
    }
    \label{fig:ui_text}
\end{figure}

\begin{figure}[!htbp]
    \centering
    \includegraphics[width=\textwidth]{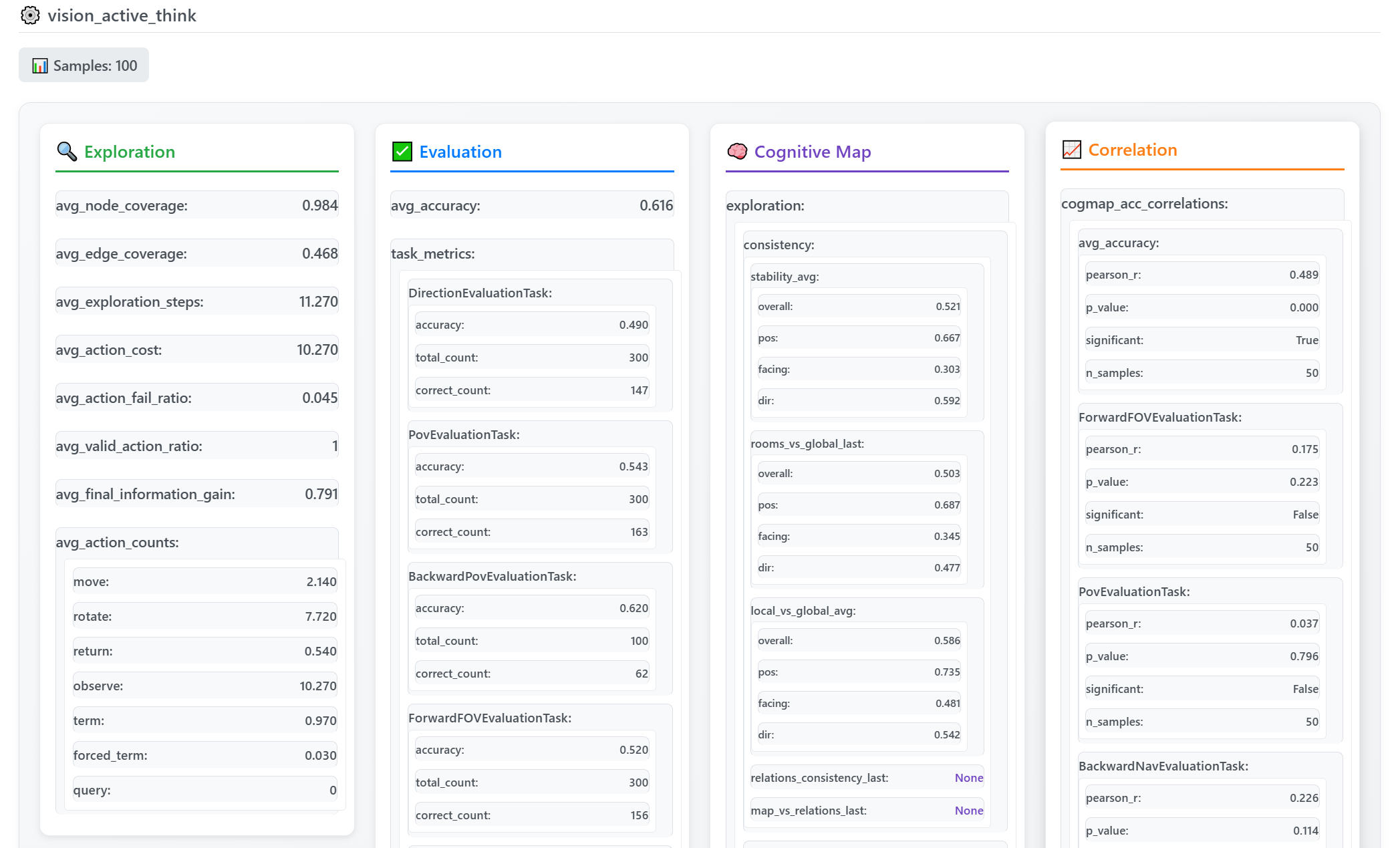}
    \caption{Visualization Platform for analysis: Metrics for active exploration in vision world
    }
    \label{fig:ui_vision}
\end{figure}

\begin{figure}[!htbp]
    \centering
    \includegraphics[width=\textwidth]{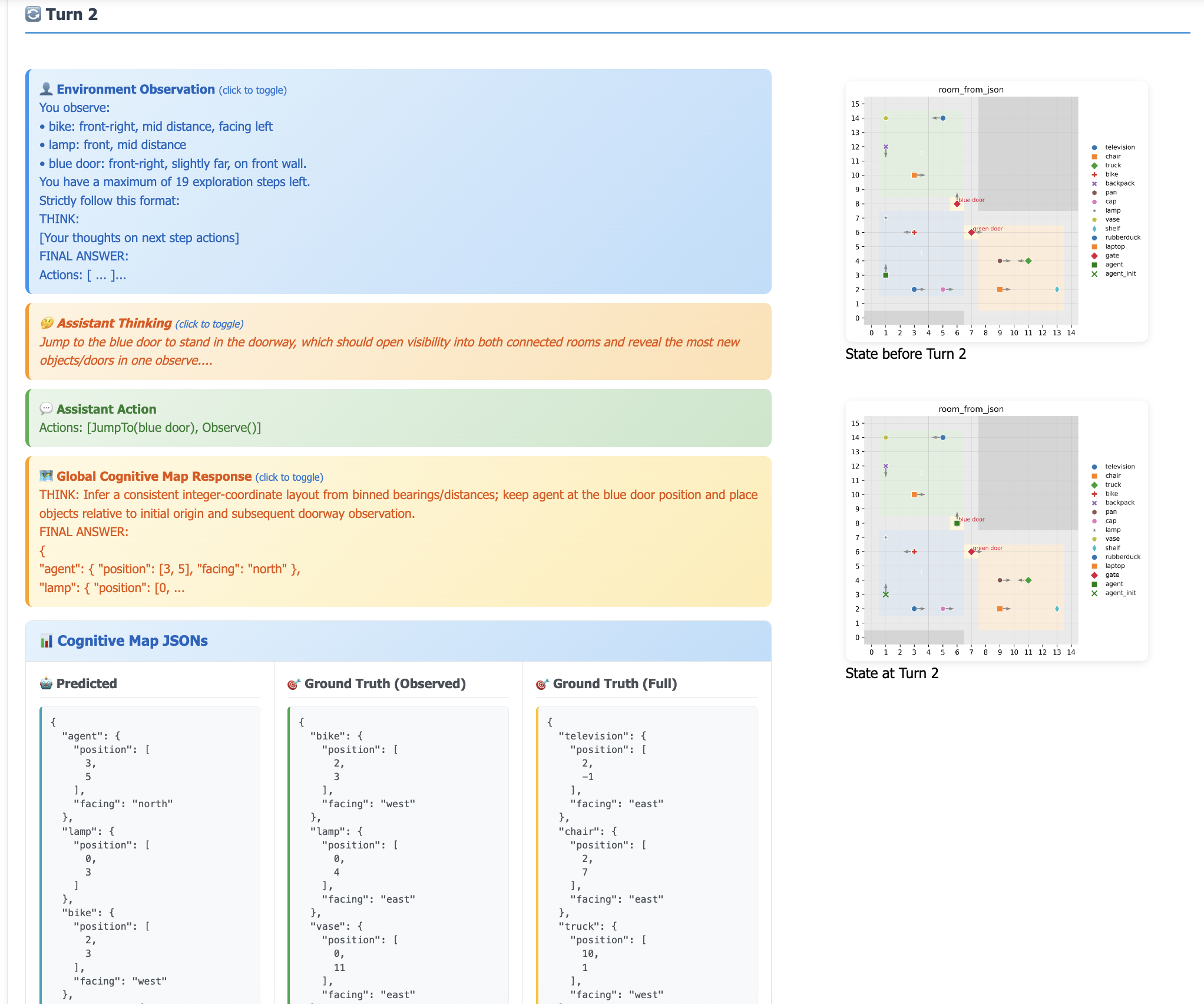}
    \caption{Visualization Platform for analysis: one turn of active exploration in text-world, including agent's action and cognitive map.
    }
    \label{fig:ui_turn_log_text}
\end{figure}

\begin{figure}[!htbp]
    \centering
    \includegraphics[width=\textwidth]{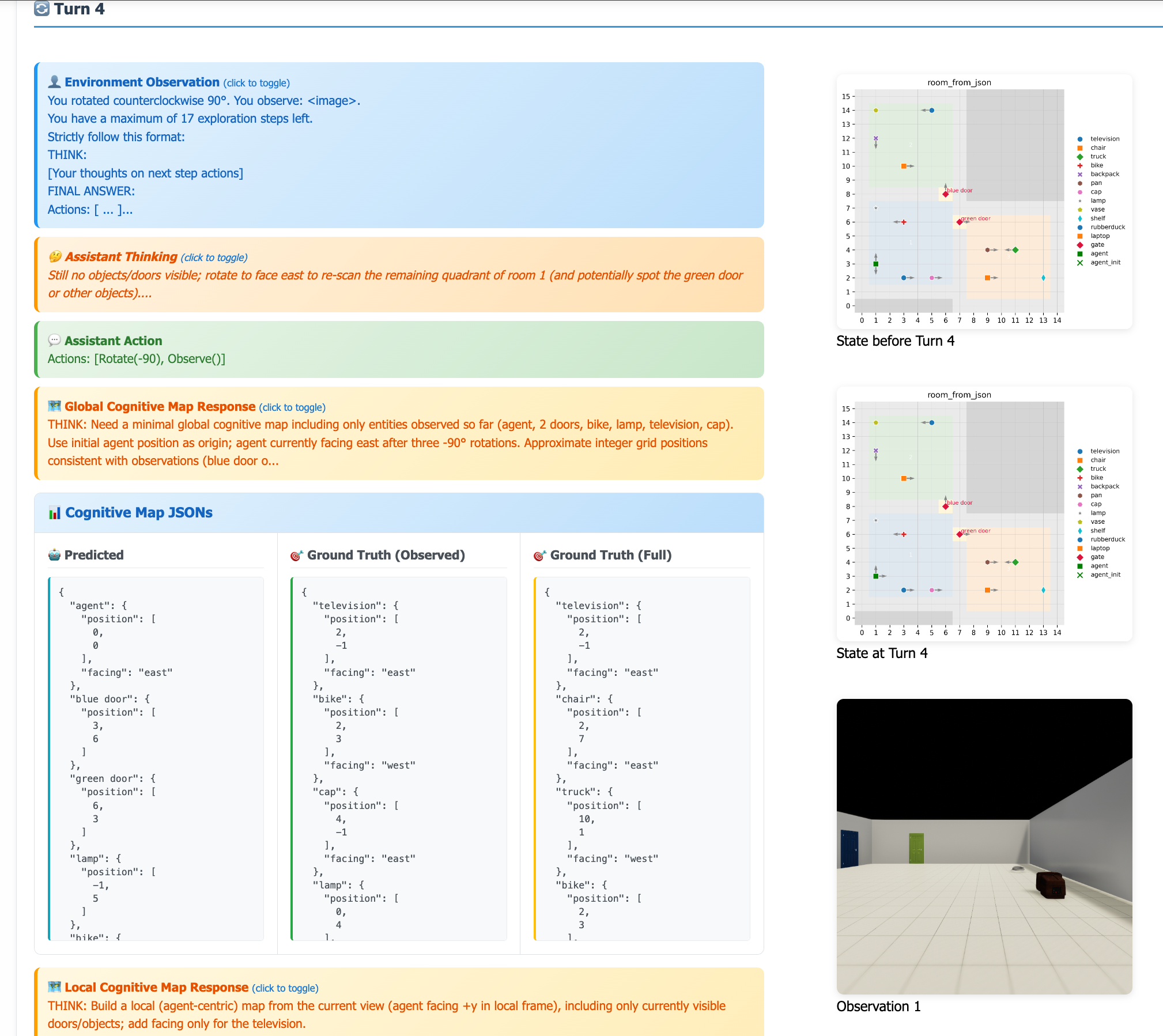}
    \caption{Visualization Platform for analysis: one turn of active exploration in vision-world}
    \label{fig:ui_turn_log_vision}
\end{figure}

\paragraph{Analysis Platform}
We also include some demonstrations in Figure~\ref{fig:ui_chart}, \ref{fig:ui_vision}, \ref{fig:ui_text}, \ref{fig:ui_turn_log_text}, and \ref{fig:ui_turn_log_vision} of our designed platform for better analysis


\end{document}